\documentclass{article}
\usepackage{arxiv}

\usepackage[utf8]{inputenc} 
\usepackage[T1]{fontenc}    
\usepackage{hyperref}       
\usepackage{url}            
\usepackage{booktabs}       
\usepackage{amsfonts}       
\usepackage{nicefrac}       
\usepackage{microtype}      
\usepackage{lipsum}

\usepackage{amsmath}
\usepackage{amsthm}
\usepackage{amssymb}

\usepackage{changepage}
\usepackage{changes}
\usepackage{tabls}
\usepackage{afterpage} 
\usepackage{placeins}

\usepackage{subcaption,siunitx,booktabs}
\usepackage{adjustbox}
\usepackage{graphicx}
\usepackage{tikz}
\usepackage{lscape}
\usepackage{longtable} 
\usepackage{blindtext}
\usepackage{tabularx}
\usetikzlibrary{shapes,arrows,shapes.geometric}

\usepackage{algorithm}
\usepackage{algorithmic}

\title{Reinforcement Learning Applied to Trading Systems: A Survey}

\author{
  Leonardo Kanashiro Felizardo \\
  Escola Politécnica\\
  Universidade de São Paulo, 
  Brazil \\
  \texttt{leonardo.felizardo@usp.br} \\
   \And
  Francisco Caio Lima Paiva  \\
  Escola Politécnica\\
  Universidade de São Paulo, 
  Brazil \\
  \texttt{francisco.paiva@usp.br} \\
   \AND
  Anna Helena Reali Costa \\
  Escola Politécnica\\
  Universidade de São Paulo, 
  Brazil \\
  \texttt{anna.reali@usp.br} \\
  \And
  Emilio Del-Moral-Hernandez \\
  Escola Politécnica\\
  Universidade de São Paulo, 
  Brazil \\
  \texttt{emilio.delmoral@usp.br} \\
}

\begin{document}
\maketitle

\begin{abstract}
Financial domain tasks, such as trading in market exchanges, are challenging and have long attracted researchers. The recent achievements and the consequent notoriety of Reinforcement Learning (RL) have also increased its adoption in trading tasks. RL uses a framework with well-established formal concepts, which raises its attractiveness in learning profitable trading strategies. However, RL use without due attention in the financial area can prevent new researchers from following standards or failing to adopt relevant conceptual guidelines. In this work, we embrace the seminal RL technical fundamentals, concepts, and recommendations to perform a unified, theoretically-grounded examination and comparison of previous research that could serve as a structuring guide for the field of study. A selection of twenty-nine articles was reviewed under our classification that considers RL's most common formulations and design patterns from a large volume of available studies. This classification allowed for precise inspection of the most relevant aspects regarding data input, preprocessing, state and action composition, adopted RL techniques, evaluation setups, and overall results. Our analysis approach organized around fundamental RL concepts allowed for a clear identification of current system design best practices, gaps that require further investigation, and promising research opportunities. Finally, this review attempts to promote the development of this field of study by facilitating researchers' commitment to standards adherence and helping them to avoid straying away from the RL constructs' firm ground.

\vspace{0.2cm}
\end{abstract}


\section{Introduction}
\label{sec:introduction}

\begin{figure}[ht]
    \centering
    \includegraphics[width=0.4\textwidth]{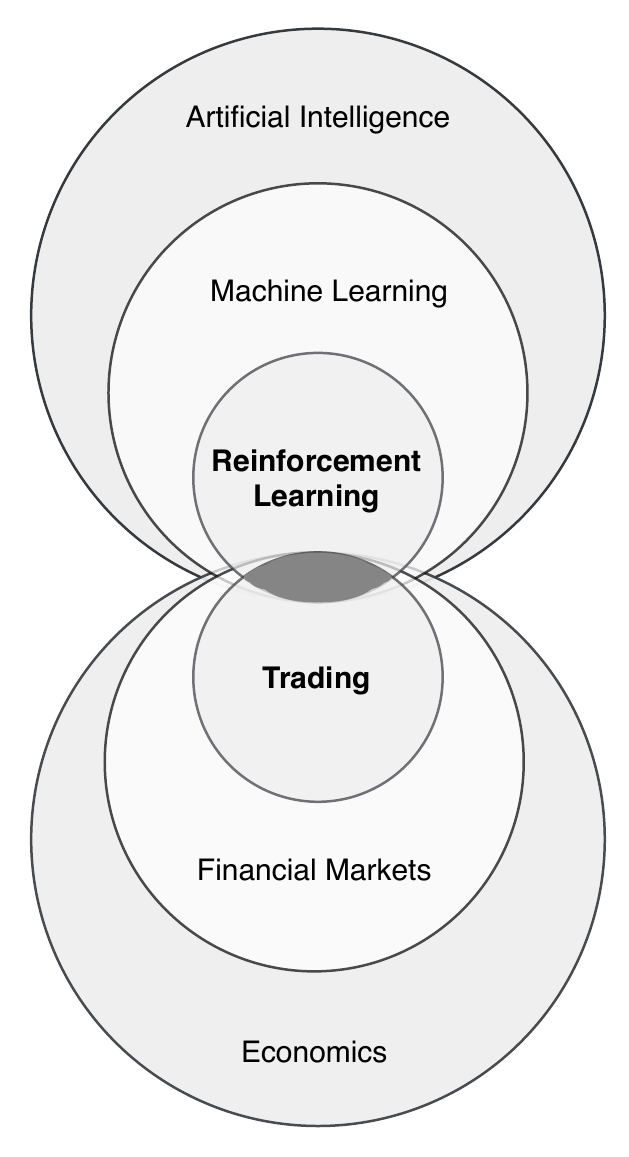}
    \caption{Interdisciplinary between Trading and Reinforcement Learning.}
    \label{fig:interdisciplinary}
\end{figure}

Financial markets perform a central role in the modern globalized economy and are a widely studied topic in economics. In particular, the subject of trading - which is a form of active engagement in financial markets - is particularly appealing to artificial intelligence (AI) researchers. Among AI approaches, machine learning is a renowned subfield that exhibits promising results at trading tasks \cite{KhadjehNassirtoussi2014, Henrique2019}. In particular, given the sequential nature of asset trading decision-making, Reinforcement Learning (RL) methods can be especially appropriate. Hence, in this work, we explore the intersection between RL and trading, as depicted in Fig. \ref{fig:interdisciplinary}.

When trading assets in the financial market, investors try to devise strategies that maximize returns, avoid losses, and attain a net profit. Some of these strategies rely on the premise that it may be possible to have at least a little foresight of the market's future behavior by identifying underlying information from market indicators. However, according to the efficient market hypothesis \cite{Fama1965} and the market's seemingly random walk behavior \cite{Cootner1967}, all the newly available information is almost instantly absorbed into the asset's prices. Thus, the market prediction is unlikely to work. Fortunately, behavioral economics supporters began an ongoing discussion over the strength of the efficient market hypothesis. From this dispute, a reconciliation line of work emerged, in which a seminal paper by \cite{Lo2004} proposed the adaptive market hypothesis. Under the general idea of the adaptive market hypothesis, not all markets are mature enough to be fully efficient, and consequently, one might be able to exploit market imperfections to make predictions and incur a profit.

Regarding machine learning for financial market tasks, some review works \cite{KhadjehNassirtoussi2014, HuYong2015, Henrique2019} showed a predominance of studies focused on the prediction of market trends using supervised learning approaches. Nevertheless, accurate prediction of market trends might not be enough to satisfy the trading task need for maximization of long-term profitability. For the trading case, estimating future returns is essential, given that the sequence of actions of an agent will affect not only the immediate return but also expected returns. In this sense, \cite{DengYue2017} showed that a simple introduction of real-world limitations, such as transaction costs, might sensibly diminish the total profitability of typical supervised learning prediction systems in the trading task scenario.

RL suitably addresses these supervised learning shortcomings \cite{DengYue2017, LiYang2019} by providing the framework for developing intelligent agents that learn trading strategies capable of maximizing accumulated expected returns. Besides, the RL framework achieves reliable performance and is adaptable to adverse conditions. This adaptability is primarily due to combining solid mathematical constructs with a flexible system's design specifications. Thus, even though RL defines properties under which convergence and stability guarantees are not necessary to devise effective agents \cite{Sutton2018}. Additionally, while the framework specifies an overall structure for sequential decision-making, this does not restrict the innovation in the design of its main elements (i.e., states, actions, rewards). For instance, RL allows customization of the rewards to fit diverse situations, and researcher's preferences \cite{Almahdi2017, Spooner2018, GiacomazziDantas2018}. Moreover, some RL methods support online learning and decision-making, a desirable characteristic for real-time trading.

Given all these advantages, there is no wonder the long history of RL adoption for different trading tasks such as portfolio optimization \cite{Neuneier1995, Neuneier1998} or single asset trading \cite{Moody1997}. Still, recent RL accomplishments (e.g., AlphaGo Systems \cite{Silver2016, Silver2017}) and improvements (e.g., development of Deep Learning RL techniques \cite{Mnih2015}) are possibly responsible for the rise in the number of works RL applications to trading problems \cite{DengYue2017, Wu2019, LiYang2019}.

This increase is noteworthy and justifies quantitative and qualitative investigations. Moreover, even though this tendency is beneficial, the subjects' interdisciplinary nature may cause researchers to overlook some fundamental RL properties. This survey aims to address these issues. We present a useful analysis from an extensive systematic review, generating insights for future work in RL and trading. The main contributions of this paper are:
\begin{itemize}
    \item A workflow {\bf pipeline} reflecting the  fundamentals of reinforcement learning  that helps to categorize and group the surveyed works for better information extraction;
    \item In-depth {\bf comparison of works} in the light of our proposed  categorization;
    \item Identification of the leading \textbf{state-of-the-art trends} in reinforcement learning efforts applied to single asset trading and portfolio optimization tasks;
    \item Insights and suggestions from the extracted information can help to \textbf{guide future work efforts}.
\end{itemize}

We explore the theoretical concepts for comprehending the literature in Section \ref{sec:foundations}. Section \ref{sec:methodology} presents the bibliometric information and article selection criteria. Section \ref{sec:review} analyzes the reviewed articles. Then, in Section \ref{sec:discussion}, there is a discussion over the analyzed information. Finally, Section \ref{sec:conclusion} concludes this work.

\section{Foundations on Trading and Reinforcement Learning}
\label{sec:foundations}

In this section, we expand on the most relevant and recurring notions observed in surveyed works. Initially, we address financial market topics and formalize the distinct types of trading problems and analysis approaches. Next, we move on to the formulation of the RL framework definitions and methods that are related to this analysis. Finally, we wrap up this section by describing the mechanisms for combining these fields of studies.

\subsection{Finance and Trading Fundamentals}
\label{subsec:finance_background}

To properly develop an RL trading system, it is fundamental to understand the main financial concepts addressed throughout this work. Therefore, in this subsection, we delve into some of the most relevant market aspects required to address trading tasks.

\subsubsection{Problem Definition: Portfolio Optimization or Single Asset Problem}

There are two main types of trading problems that an RL solution can address, and their differences entail distinct system designs. Given the number of traded assets, we have the following types:

\textbf{Single asset problem} is defined by the trading of one single asset at each operation during a given period. The agent can perform actions that affect the asset's position, such as buy, sell, and hold. In some scenarios, when the agent can borrow assets, we represent the actions by the positions long, short, and neutral, usually encoded by either the set of discrete $A=\lbrace-1,0,1\rbrace$ or continuous  $A=[-1, 1] \in \mathbb{R}$ actions. 

\textbf{Portfolio optimization}, also known as a portfolio management problem, encompasses the primary purpose of optimizing the allocation of multiple assets with pre-established constraints such as total portfolio amount and number of assets per type (e.g., fixed income, stocks, derivatives, and others). The agent can only buy or sell the assets individually to allocate different amounts of each one. Hence, the portfolio optimization objective is to maximize the portfolio value over time and minimize the risk (usually associated with the volatility of the portfolio value). In this sense, the agent takes actions that involve a change in the amount of each asset in the portfolio, represented by a vector of weights $w = (w_1,w_2,...,w_n)$, where $w_i$ is the weight of the $i$ asset of the portfolio comprising $n$ assets. Naturally, the agent can change the vector of weights to maximize the portfolio's future value on each decision.

Independent of the type of problem, either for single asset trading or portfolio optimization, the agent can observe a feature space composed of prices, technical indicators, or fundamental indicators. These different forms to represent the observation space relate to a classical point of view divergence in the financial economics literature.

\subsubsection{Fundamental analysis versus technical analysis}
\label{subsec:types_of_market_analysis}

Among investors that assume some degree of market exploitability, the predominant viewpoints include the combination of two approaches: technical and fundamental analysis. The difference between these two groups starts with assumptions about what type of financial indicators contain more relevant information.

\textbf{Technical analysis} finds information present in the assets' price time-series to be sufficient for operation. Therefore, this approach employs either unprocessed (e.g., crude price time-series values) or preprocessed (e.g., using moving averages, relative strength indexes, stochastic oscillators, and other techniques) technical indicators. The price time-series typically comprises four primary assets information at each given time-frame (i.e., period): prices of the asset at the beginning (i.e., open) and end of the period (i.e., close); assets' highest (i.e., high) and lowest observed prices in the period (i.e., low). Additionally, although the popular \textbf{OHLC} data acronym stands for open, high, low, and close prices, it is usual for this data to contain information about the total negotiated volume of an asset during a period. To avoid correlation in this sequential data, authors typically employ the price difference $z$ between sequences of price data points. 

\textbf{Fundamental analysis} disputes the assumption that all relevant information is unequivocally present in the price time-series alone and, thus, encourages the adoption of a much broader set of available information. To evaluate profitability potential, this type of analysis regularly uses fundamental indicators (e.g., price to cash flow, profit and losses, price to earnings, and others) extracted from companies' periodical financial reports. Therefore, fundamental analysts are usually less restrictive about their sources of inputs and may also consider macro-economic indicators, overall industry numbers, financial news, and others.

Alternatively, it is also possible to employ both types of indicators and execute a hybrid approach. For instance, some machine learning review works \cite{KhadjehNassirtoussi2014, HuYong2015, Henrique2019} already observed a reasonable amount of studies that combined indicators from both types of analysis. The features generated from the technical or fundamental analysis compose the agent's observation space in the reinforcement learning framework, further described ahead.

\subsection{Basic reinforcement learning concepts}

Reinforcement learning is a broad concept that, at the same time, refers to a problem, a set of adequate methods to solve these problems, and the field of study encompassing them \cite{Sutton2018}. As a problem, it describes situations where an agent needs to map states to actions for numerical reward signal maximization. Also, in RL, there is no supervisor; the only feedback is the reward. This feedback may be delayed, and since RL is a sequential decision process, it is essential to account for correlation in the data. Moreover, since the situations are auto-correlated, we may not interpret it as a classical supervised learning problem where data is \textit{independent and identically distributed}. As an additional contribution, we present a notation that unifies observed representations among surveyed articles around standard symbols, nomenclature, and formulations of the seminal RL literature \cite{Sutton2018}.
 
\begin{figure}[ht]
    \centering
    \includegraphics[width=0.6\linewidth]{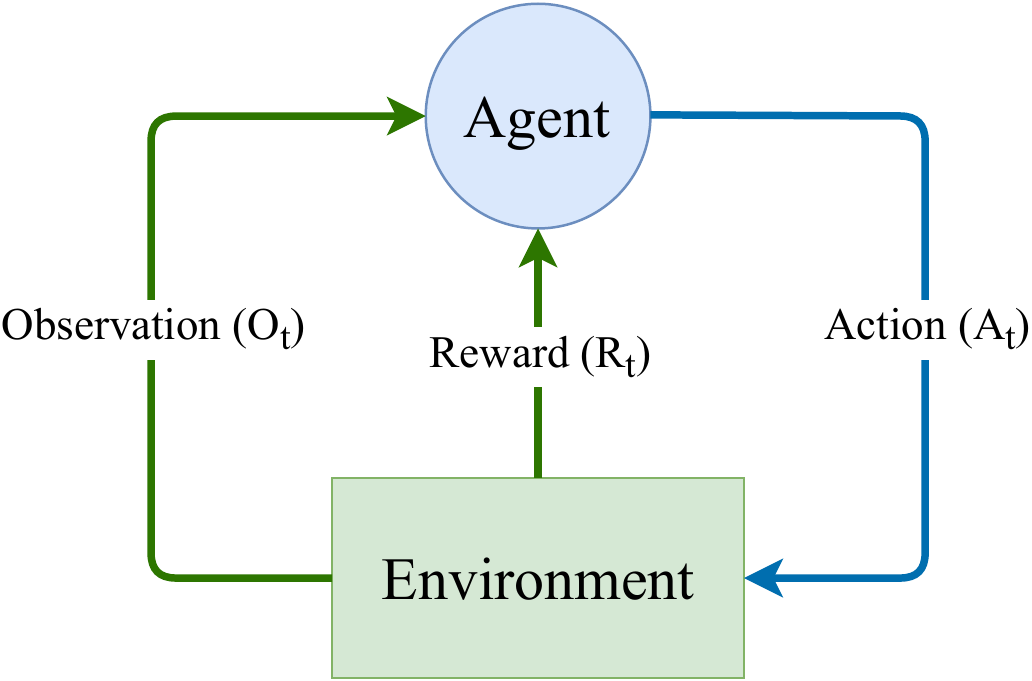}
    \caption{Reinforcement learning framework. Interaction of the agent through action $A_t$, receiving the environment feedback with observation $O_t$ and reward $R_t$.}
    \label{fig:rlframework}
\end{figure}

In the reinforcement learning framework, the agent interacts with the environment at step $t$ as in Figure \ref{fig:rlframework}, executing action $A_t$, receiving observations $O_t$ and earning the scalar reward $R_r$. Each action the agent performs may influence the environment, affecting future rewards.

The sequence of actions, observations and rewards is called history, $H_t = A_1, O_1, R_1, ..., A_t, O_t, R_t$. The history will determine the agent's next decision and what is the environment's next observation and reward. The agent uses a summary of the information present in history to determine the future. The agent also uses the information history represented as the state defined as a function of the history, $S_t = f(H_t)$. 

Ultimately, to solve the RL problem, the agent's objective is to find an adequate mapping of states and actions that lead to good outcomes. A policy $\pi$ is a mapping that determines (or gets sufficiently close to determining) the optimal behavior $\pi^*$ to a given task. To reach an adequate policy, researchers can adopt methods that either estimate value functions, approximate a policy directly, or both. For example, the value function $V(S_t)$ helps measure the usefulness of the agent being in a specific state, given that it continues following a policy $\pi$.

\subsubsection{Contextual bandit and full reinforcement learning}
\label{subsec:contextual_bandit}

There are distinct ways to represent an RL problem. In their book, \cite{Sutton2018} presents the concept of a \textit{contextual bandit}\footnote{Bandit is equivalent to a slot machine in casinos.} problem where tasks are associative, implying that different actions may lead to different outcomes depending on the states. However, according to \cite{Sutton2018}, a contextual bandit problem is a special case of the reinforcement learning problem because of the simplification that an agent's action can exclusively influence its immediate reward. Thus, the contextual bandit can assume a representation as the tuple \(\langle \mathcal{S,A,R} \rangle\). Following this representation, we have that an agent observes a given state $S_t \in \mathcal{S}$ at each discrete time-step $t=0,1,2,3,...$, takes an action $A_t \in \mathcal{A}$ and, as a consequence of each action, receives the reward $R_t \in \mathcal{R}$, $\mathcal{R}=\mathbb{E}[R_t+1|S_t=s,A_t=a]$.

\begin{figure}[ht]
    \centering
    \includegraphics[width=0.8\linewidth]{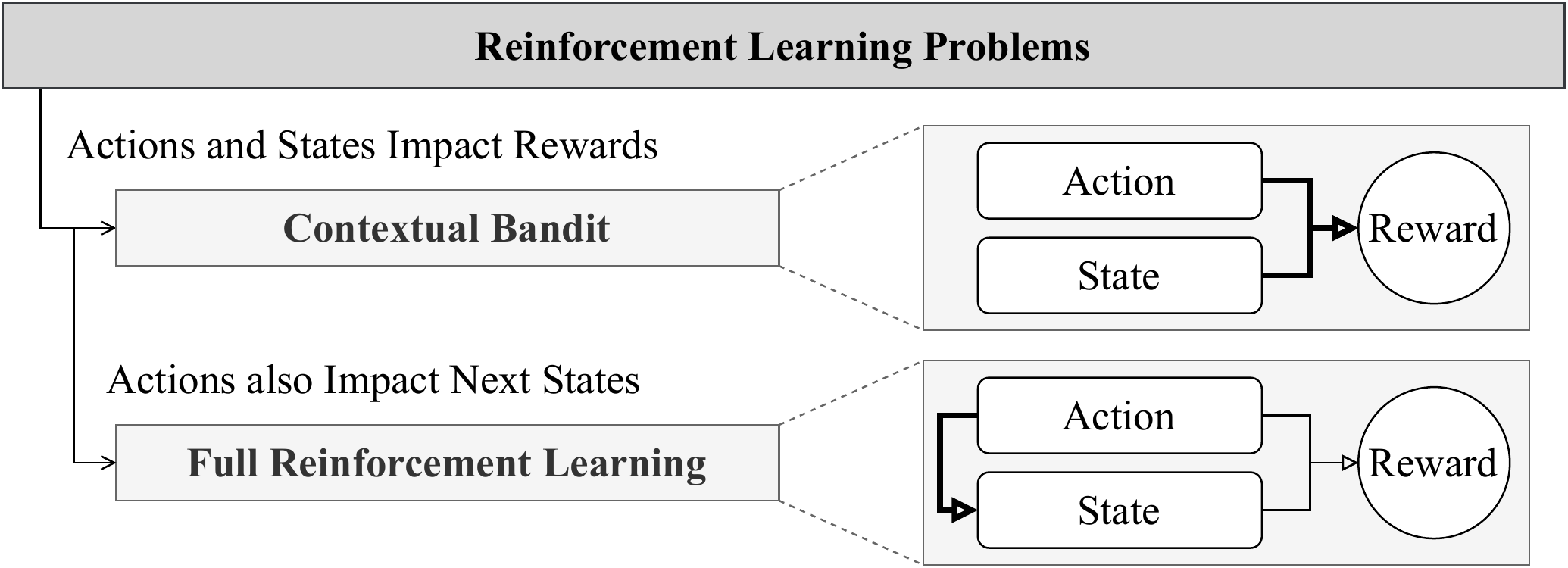}
    \caption{Comparison between the contextual bandit and full reinforcement learning types of problems.}
    \label{fig:cb-vs-rl}
\end{figure}

On the other hand, in a \textit{full reinforcement learning} problem, each action has to impact both the reward and the next state. In Fig. \ref{fig:cb-vs-rl}, we have a visual summary of these classification differences regarding RL problems.

\subsubsection{Markov decision process formulation}

It is usual for full reinforcement learning problems to adopt the Markov decision process (MDP) formulation. MDPs solve problems involving evaluative feedback, considering agents interacting with an environment (everything outside the agent) to achieve a goal \cite{Sutton2018}. In the MDP, the environment has the Markov property if and only if the state $S_t$ captures all the  necessary information from the history, $\mathbb{P}[S_{t+1}|S_t] = \mathbb{P}[S_{t+1}|S_1, ...,S_t]$. Sometimes, a researcher can convert partially observable problems into MDPs. The Markov decision process happens in an environment with the Markov property. Ultimately, we have a random sequence of states that follow Markov's property, in which the future states depend only on the present state and not on the past states, defining a Markov process.

For the Markov decision process, we have a tuple $\langle \mathcal{S,A,P,R},\gamma \rangle$. This representation is similar to the contextual bandit formulation with the addition of the state-transition probabilities $\mathcal{P} = \mathbb{P}[S_{t+1}|S_t=s, A_t=a]$ and the discount factor $\gamma\in[0,1]$ for calculating the accumulated return $G_t = R_{t+1} + \gamma R_{t+2}+...=\sum^\infty_{k=0} \gamma^k R_{t+k+1}$. The probabilities \(\mathcal{P}\) describe the agent's model (i.e., acquired knowledge of the environment dynamics). On a different note, the purpose of the discount factor \(\gamma\) is to account for the uncertainty about the future, avoid infinite returns in cyclic Markov processes, and to mathematical convenience.

\subsection{Reinforcement learning applied to the trading problem}
\label{subsec:applied}

Reinforcement Learning methods are notorious for their successes in the game domain by solving both board \cite{Silver2016, Silver2017}, and video game \cite{Mnih2015} tasks. The logic behind games presumes a scenario with a well-defined state representation and situations where the actions directly impact the following state and reward. Thus, using a full RL formulation seems almost innate for these scenarios. However, for some domains, such as the financial market here considered, the definition of the state representation is not as straightforward and requires careful identification of relevant features. Consequently, depending on which features RL approaches on the use of the financial domain to represent their state, they may be classified as either a contextual bandit or a full reinforcement learning formulation.

These definitions may seem rather conceptual, but \cite{Sutton2018} examined two works on the web marketing domain \cite{LihongLi2010, Theocharous2015} to emphasize that these formulations have practical implications. These implications include the design of the state representation and the method that theoretically is most efficient for solving that RL formulation. For example, \cite{LihongLi2010} used a contextual bandit formulation and designed its states accordingly. \cite{LihongLi2010} then proposed a variation of the upper confidence bound (UCB) algorithm, which is popular in solving bandit problems. \cite{Theocharous2015}, on the other hand, aimed at a full RL formulation and compared the proposed RL method against a contextual bandit method. This comparison showed the practical differences between the choice of formulation and the method's consequent choice for solving the problem. In Section \ref{subsec:preprocessing}, we use these concepts just described to classify works, analyze their particularities, and discuss the practical implications of this classification.

Furthermore, the MDP formulation is a conventional representation that, for trading tasks, can favor the representation of states as a composition of time-series of prices or returns. This representation could have many implications on the Markov property's validity since the price time-series is insufficient to explain all the future states, which characterizes a partial observability scenario. As a usual approximation, the authors assume that the price time-series themselves suffice to overcome the partial observability problem. Alternatively, researchers may assume other approximations to satisfy the Markov property and fit a task as an RL problem. All these different assumptions can affect the choice of the proper RL method. In Section \ref{sec:discussion}, we resume this observability subject, and in the following subsection, we analyze different types of RL methods.

\subsubsection{Temporal difference methods}

Temporal Difference (TD) methods move toward the globally optimal policy by updating value function estimates with the difference in consecutive states' values. TD uses the state-action value function \(Q(S_t, A_t)\), also known as Q-value, to evaluate the state and action tuple at each step. This value function guides the agent toward the maximization of the expected reward. When using the deep Q-networks, approximates the value function with a neural network \cite{Mnih2015}. Also, by \textit{bootstrapping} \(Q(S_t,A_t)\) initial values, TD-learning agents can learn from raw experience without also learning the dynamics \(\mathcal{P}\) (\textit{model-free} RL) of the environment \cite{Sutton2018}. Using these functions to estimate and direct the agent to the optimal policy \(\pi^*\) contributed to adopting the term \textit{critic} to refer to these value-based methods.

For the first application of the TD approach on portfolio optimization, \cite{Neuneier1995} applied the Q-learning method. In Q-learning, the agent learns the optimal value function \(Q^*(S_t, A_t)\) by sampling state-action pairs and returns from its interaction with the environment. This procedure recursively updates the Q-value according to the equation \ref{eq:q-learning-neuneier},
\begin{equation}
Q(S_t,A_t) \leftarrow Q(S_t,A_t)+ \alpha [R_{t+1}+\gamma \max_{a}Q(S_{t+1},a) - Q(S_t,A_t)],
\label{eq:q-learning-neuneier}
\end{equation}
The agent executes an action at state $S_t$ moving to state $S_{t+1}$, which, for the portfolio optimization case, numerous price time-series compose the state $S$. Where $ \alpha $ is the learning rate, and $\gamma$ is the discount factor. The discounted Q-value is the maximum Q-value among all actions in the action space of the subsequent state.

Another TD-learning approach for the trading task \cite{Pendharkar2018,Spooner2018,Alimoradi2018} is the SARSA method \cite{Rummery94}. Contrary to Q-learning, SARSA is an \textit{on-policy} control algorithm that estimates the value function \(Q(S_t,A_t)\) while following the current policy's behavior. Alternatively, \textit{off-policy} methods, such as Q-learning, may use actions that are not determined by current policy to adjust the value function estimates.

The SARSA algorithm uses the quintuple of states and actions, $(S_t,A_t,R_{t+1},S_{t+1},A_{t+1})$, 
which originates from the acronym for SARSA, and 
\begin{equation}
Q(S_t,A_t) \leftarrow Q(S_t,A_t)+ \alpha [R_{t+1}+\gamma Q(S_{t+1},A_{t+1}) - Q(S_t,A_t)] 
\end{equation}
as $Q$ update equation.

Assuming that researchers adequately formulate the problem, TD learning can reach the optimal value function, which determines the optimal policy. However, the value function approach, usually implemented with TD-learning methods, does not work for continuous action spaces. Therefore, some authors work with discrete action spaces such as $A=\lbrace-1,0,1\rbrace$. Direct policy approximation methods are an alternative to value function evaluation, and they can address this continuous action space issue.

\subsubsection{Policy approximation methods}

Given that the policy approximation methods learn the policy \(\pi\) directly, without using a value function, they are also commonly referred to as \textit{actor} approaches. Policy gradient is a conventional method for direct approximation of policies, and even though they might not lead to an optimal solution, they are quite popular \cite{Sutton2018}. Thus, they are expected to be a favored method for trading applications \cite{DengYue2017,Wu2019,Aboussalah2020}. The gradient method employs policy parameter learning based on a $J(\theta) $ performance measure. These methods maximize the performance $J(\theta)$ by updating the policy with approximated gradient ascent:

\begin{equation}
\theta_{t+1} = \theta_{t}+\alpha \nabla J(\theta_{t}), 
\label{eq:01}
\end{equation}
where we have the model parameters$\theta$, the gradient over the performance measure $\nabla J(\theta_{t})$, and the learning rate $\alpha$.

We consider recurrent reinforcement learning (RRL) as a policy gradient method, and we will further discuss this categorization in Section \ref{subsec:rl_methods}. As we observe further, policy gradient-based methods appear in almost half of the last five years' publications. Notably, RRL is the most employed technique among the reviewed papers. One other variant of the policy gradient used by one of the reviewed authors \cite{DingYi2018} is the REINFORCE, the policy gradient with Monte Carlo rollouts to compute the rewards.

\subsubsection{Actor-critic methods}

Actor-critic methods are a hybrid combination of previously discussed RL methods, in which the agent directly approximates a policy (actor) and then estimates its value function (critic). Thus, the value function is used as a baseline for the policy approximation to avoid the high variance that the gradients can reach. This baseline can have different implementations depending on the technique, such as advantage actor-critic (A2C) \cite{Baird1993} or the Q actor-critic\cite{Sutton2000}. More recently, a variation of the A2C emerged, the asynchronous advantage actor-critic (A3C) \cite{Mnih2016}, that employs multiple agents to interact with different instances of the environment in an asynchronous form that accelerates the exploration. To calculate the performance function for an advantage actor-critic, we have
\begin{equation}
\nabla_\theta J(\theta) = \mathbb{E}_{\pi_\theta}([\nabla_\theta log \pi_\theta(s,a)A(s,a)]),
\end{equation}
where $A$ is the advantage function which is calculated by
\begin{equation}
A(s_t,a_t) = Q(s_t,a_t) - V(s_t),
\end{equation}
that can also be rewritten as 
\begin{equation}
A(s_t,a_t) = r_{t+1} + \gamma V(s_{t+1}) - V(s_t).
\end{equation}

There is a variant of the A2C, called deep deterministic policy gradient (DDPG) \cite{Lillicrap2016} that can also operate in the trading domain \cite{Yu2019,Ye2020}. The fundamental characteristic of the DDPG method is that the actor learns by taking deterministic actions instead of the sampling actions from a probability distribution and still can achieve good performance if adequately implemented \cite{Silver2014}. Additionally, some authors adopted R-learning \cite{Schwartz1993}, another variant of the actor-critic approach. The R-learning method seeks to maximize the average reward per time step. In this case, the value function for a policy employs the average expected reward \cite{Sutton2018} and, therefore, does not require the discount factor.

\subsubsection{The exploration–exploitation dilemma}
\label{subsec:exploration_exploitation}

The exploration-exploitation dilemma is a central subject for the reinforcement learning field. When exploring an unfamiliar environment, the RL agent will identify that, at each time step, there could be one \textit{greedy} action that leads to a higher immediate reward than other actions. In this sense, exploitation happens when one agent prioritizes selecting greedy actions. Alternatively, exploration occurs when an agent experiments with taking non-greedy actions. One of the RL challenges is to balance the trade-off between exploiting existing behavior and exploring an alternative, possibly, a better course of action \cite{Sutton2018}.

Achieving an optimal balance between exploitation and exploration is still an unsolved problem. Fortunately, there are mechanisms for each type of RL approach that help to address this issue. For example, temporal difference methods can resort to the \(\varepsilon\)-greedy mechanism, where an agent chooses a non-greedy action with a probability \(\varepsilon\). Policy approximation methods can increase stochasticity in selecting an action by sampling actions from a probability distribution. In Section \ref{subsec:rl_methods}, we discuss how different methods address the exploration-exploitation dilemma.

\section{Selection of the articles for the review}
\label{sec:methodology}

From the analysis of previous surveys about the adoption of machine learning for financial problems \cite{KhadjehNassirtoussi2014,HuYong2015,Henrique2019}, we recognized a particular tendency that motivated our present work. Most articles surveyed in these works adopted methods with static behaviors given by predefined rules over the prediction of supervised learning models (e.g., buying an asset when the predicted trend is positive). This prevalence occurs despite some researchers \cite{Neuneier1995,Moody1997} paving the way for strategy learning via reinforcement learning. Also, \cite{HuYong2015} pointed to the importance of online training for real-world trading, which is a straightforward mechanism for some RL methods.

As a result, we work with two central assumptions. The first is that learning trading strategies fit a less explored path than predetermined rule strategies, even if the latter is based on machine learning methods. The second is that the RL framework is a more suitable machine learning approach for learning trading strategies.  

\begin{figure}[htb]
\centering
\includegraphics[width=1\linewidth]{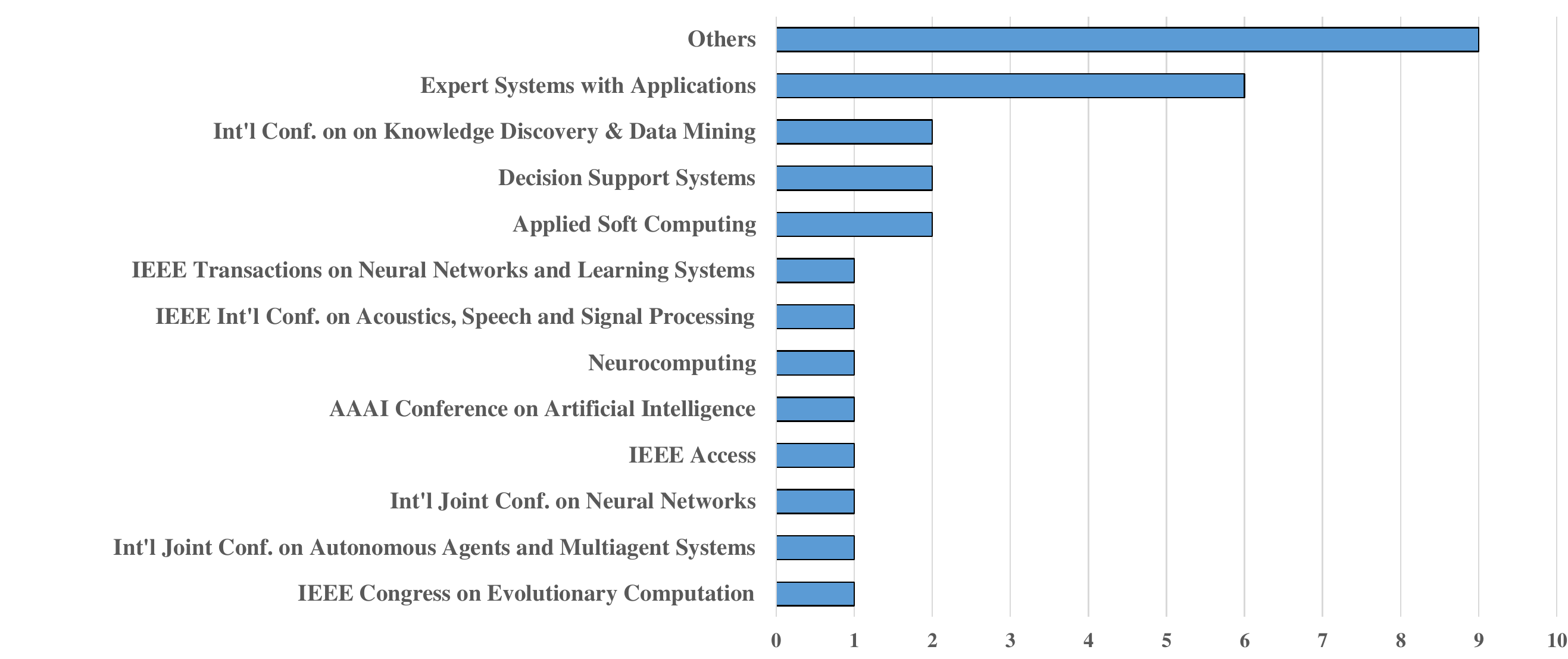}
\caption{Distribution of articles over different journals and conferences.}
\label{fig:dist_conf_jour}
\end{figure}

For the selection of works, we established three filtering criteria. First, the article's primary task must be to approach the single asset trading problem or the portfolio optimization problem in the financial market. Consequently, we discarded exclusively predictive works. Second, articles should focus on learning trading strategies, and thus, we discarded articles that used predetermined rules based on financial indicators or machine learning prediction systems. Finally, works must primarily use the reinforcement learning framework to learn those strategies. Given these criteria, Fig. \ref{fig:dist_conf_jour} depicts the distribution of the selected papers over the publication vehicles.

The list of selected articles, with each respective publication vehicle, is presented in Table \ref{tab:articles-comp}. Also, given the plurality of observed publication vehicles and that many received only one of the reviewed articles, we group some of them in the category ``others".

\begin{figure}[htb]
\centering
\includegraphics[width=0.8\linewidth]{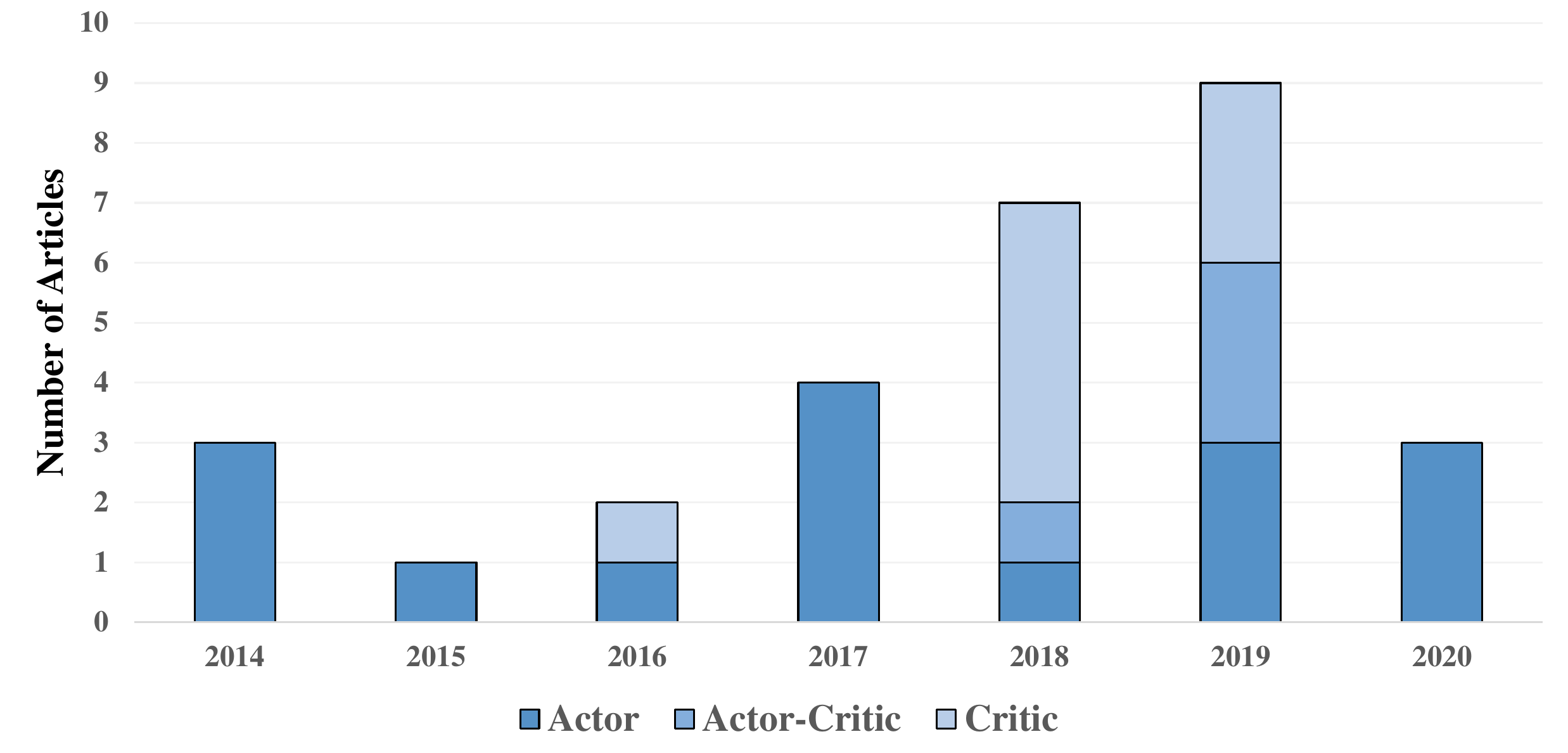}
\caption{Distribution of the RL techniques classes, actor, critic and actor-critic, over the last seven years.}
\label{fig:disttecyear}
\end{figure}

To overview the application of reinforcement learning to the trading problem and extract highlights from authors' different approaches, we evaluated articles from the year 2014 up to 2020. Using appropriate keywords such as ``trading", ``reinforcement learning," and others, we gathered 29 articles following our previously described selection criteria. We extracted the articles from standard database resources (e.g., Scopus, IEEE Xplorer, Google Scholar, and others).
Fig. \ref{fig:disttecyear} shows the distribution of RL techniques over the years and presents the respective categories of actor, critic, and actor-critic.

We can observe in Fig. \ref{fig:disttecyear} an increase in published works of interest after 2016. Recent advancements in the RL field might help to explain this tendency. For instance, some studies expanded the field's notoriety by creating the superhuman performing system AlphaGo, for the Go board game \cite{Silver2016, Silver2017}. Also, novel developments of deep learning permeated the appearance of deep reinforcement learning techniques \cite{Mnih2015}, which helped to address the curse of dimensionality issues in RL \cite{Sutton2018}. Thus, the advent of deep RL serves to adjust actions more precisely based on previously observed conditions, increasing the agent's history-retaining capacity.

\begin{figure}[htb]
\centering
\includegraphics[width=0.8\linewidth]{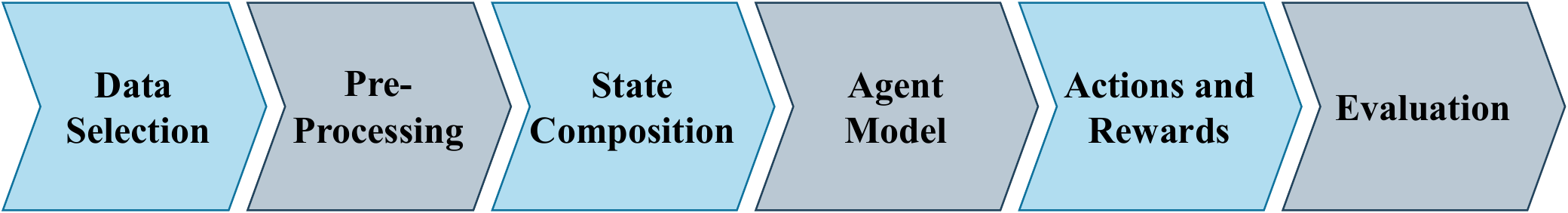}
\caption{RL pipeline workflow for single asset trading or portfolio optimization problem.}
\label{fig:pipeline}
\end{figure}

Ultimately, we propose an analysis organization that reflects the typical steps authors might go through to build an RL system. We depicted this workflow pipeline in Fig. \ref{fig:pipeline}, and it is composed of six sequential steps for a proper RL method execution and performance assessment.

\noindent\begin{minipage}{\textwidth}
{\small
\setlength{\LTleft}{-20cm plus -1fill}
\setlength{\LTright}{\LTleft}
\begin{longtable}{@{} p{0.24\textwidth} p{0.50\textwidth} p{0.35\textwidth} @{} }
\caption{Reviewed articles and the respective publication vehicle.}
\label{tab:articles-comp}\\
\toprule
\textbf{Reference} &\textbf{Publication Vehicle} \\ \hline
\endhead
\cite{Aboussalah2020} & Expert Systems with Applications \\ \hline
\cite{Weng2020} & Neurocomputing \\ \hline
\cite{Ye2020} & AAAI Conference on Artificial Intelligence \\ \hline
\cite{Park2019}  & Expert Systems with Applications \\ \hline
\cite{Almahdi2019} & Expert Systems with Applications \\ \hline
\cite{Jeong2019}  & Expert Systems with Applications \\ \hline
\cite{LiYang2019} & IEEE Access \\ \hline
\cite{Ponomarev2019} & Journal of Communications Technology and Electronics \\ \hline
\cite{Wang2019a}  & Int'l Conf. on on Knowledge Discovery \& Data Mining\\ \hline
\cite{Wu2019} & Int'l Joint Conf. on Neural Networks \\ \hline
\cite{Yu2019}  & Arxiv \\ \hline
\cite{Zarkias2019} & Int'l Joint Conf. on Autonomous Agents and Multiagent Systems \\ \hline
\cite{Alimoradi2018}  & Applied Soft Computing \\ \hline
\cite{Carapuco2018} & Applied Soft Computing \\\hline
\cite{DingYi2018} & Int'l Conf. on on Knowledge Discovery \& Data Mining\\ \hline
\cite{GiacomazziDantas2018} & Brazilian Conf. on Intelligent Systems \\ \hline
\cite{Kang2018} & Int'l Conf. on Big Data Research \\ \hline
\cite{Pendharkar2018} & Expert Systems with Applications \\ \hline
\cite{Spooner2018} & Int'l Joint Conf. on Autonomous Agents and Multiagent Systems \\ \hline
\cite{Almahdi2017} & Expert Systems with Applications \\ \hline
\cite{DengYue2017} & IEEE Transactions on Neural Networks and Learning Systems \\ \hline
\cite{JiangZhengyao2017} & Intelligent Systems Conference \\\hline
\cite{Si2017} & Int'l Symposium on Computational Intelligence and Design \\ \hline
\cite{Feuerriegel2016} & Decision Support Systems \\ \hline
\cite{Zhang2016} & Computational Economics \\\hline
\cite{Gabrielsson2015} & Int'l Symposium on Computational Intelligence and Design \\ \hline
\cite{Eilers2014} & Decision Support Systems \\\hline
\cite{Maringer2014} & IEEE/IAFE Conference on Computational Intelligence for Financial Engineering \\ \hline
\cite{ZhangJin2014} & IEEE Congress on Evolutionary Computation\\ \bottomrule
\end{longtable}}
\end{minipage}

\section{Review of the selected literature}
\label{sec:review}

We employ the pipeline proposed in Section \ref{sec:methodology} to outline the present section and organize the subsequent analysis. Although we make some minor remarks in this section, we aimed to execute a general comparative study of systems design among articles. Hence, we postpone overall suggestions and conclusions to later Sections.

\subsection{Raw data source}
\label{subsec:raw_data}

We start our analysis by separating the surveyed works according to the financial market trading problems they were addressing. We consider that this work division among types of problems or tasks, is a notable characteristic that affects all design decisions, and so we emphasized this grouping in both Tables \ref{tab:data_1}, \ref{tab:data_2}, \ref{tab:data_3}, and \ref{tab:data_4}. Additionally, about forty percent of the surveyed papers considered the portfolio optimization task with the agent balancing assets in a wallet. On the other hand, most of the authors opted to evaluate their systems' performance on each asset, as a single asset trading task.
\noindent\begin{minipage}{\textwidth}
{\small
\setlength{\LTleft}{-20cm plus -1fill}
\setlength{\LTright}{\LTleft}
\begin{longtable}{@{} p{0.28\textwidth} p{0.07\textwidth} p{0.1\textwidth} p{0.47\textwidth}  @{} }
\caption{Comparison of input data according, type of asset, the origin country of each asset, and a brief description of relevant information for the portfolio optimization problem.}\label{tab:data_1}\\
\toprule
\textbf{Portfolio optimization}\\
\toprule
\textbf{Reference} & \textbf{Type} & \textbf{Data from} & \textbf{Description} \\ \hline
\cite{Weng2020} & FOREX & U.S & 20 cryptocurrencies\\ \hline
\cite{Aboussalah2020} & Stock & U.S & 10 selected stocks\\ \hline
\cite{Ye2020} & FOREX, Stocks& U.S & 10 cryptocurrencies and 9 hightech companies \\ \hline
\cite{Park2019} & Stocks & U.S., Korea & 3 U.S. ETFs index: S\&P500 index, Russell 1000 inndex, Russell Microcap index and \newline 3 Korean ETFs index: KOSPI 100 index, KOPSI midcap index, KOPSI microcap index \\ \hline
\cite{Almahdi2019} & Stocks & U.S. & All S\&P 100 stocks \\ \hline
\cite{Wang2019a} & Stocks & China, U.S. & 1000 stocks (U.S.), 1131 A-share stocks (China) \\ \hline
\cite{Yu2019} & Stocks, Indexes & U.S. &  Stocks: Costco, Cisco, Ford, Goldman Sachs, AIG and Caterpillar; Index: S\&P 500 \\ \hline
\cite{Pendharkar2018} & ETFs, Bonds & U.S. & ETF: index-based on S\&P 500: Aggregate Bond Index (AGG) or a 10-year U.S. T-note \\ \hline
\cite{Kang2018} & Stocks & U.S. & 50 stocks without incomplete data \\ \hline
\cite{DingYi2018} & Stocks & China & 167 stocks from HS 300, including Oil, Steel, Retail, Bank, Electronic and others \\ \hline
\cite{Almahdi2017} & ETFs & U.S. & iShares Russell 1000 Value (IWD), iShares Micro-Cap (IWC), SPDR S\&P 500 ETF (SPY), WisdomTree Emerging Markets High Dividend, iShares 10+ Year Credit Bond (CLY) \\ \hline
\cite{JiangZhengyao2017} & FOREX & U.S. & 12 highest volume cryptocurrencies from Poloniex exchange \\ \hline
\cite{Zhang2016} & Stocks & U.S. & 180 stocks from S\&P 500 without incomplete data \\
\bottomrule
\end{longtable}}
\end{minipage}

The selected articles employed input data comprising price-points of different assets, from various countries and other characteristics, as observed in Tables \ref{tab:data_1},\ref{tab:data_2},\ref{tab:data_3} and \ref{tab:data_4}. Among types, the use of stocks is prevalent. However, about a third of the reviewed works opted to use stock market indexes data either directly \cite{Eilers2014, Yu2019, Ponomarev2019, Jeong2019} or indirectly through proxies such as exchange-traded funds (ETFs) \cite{Almahdi2017, Pendharkar2018} and futures contracts \cite{Gabrielsson2015, Si2017, DengYue2017, LiYang2019}. Other works opted to use the FOREX (i.e., trading currencies) data for Euro, U.S. dollars \cite{Carapuco2018}, or even cryptocurrencies \cite{JiangZhengyao2017}. One advantage of cryptocurrencies over more conventional types of assets is that their trading occurs on markets that operate 24/7 instead of only during working hours, which enables the collection of higher amounts of data.

We are considering the number of assets selected for data collection. Most works gathered data for twenty assets. On the other hand, it is usual for a system to choose some assets among a pool of assets for portfolio optimization problems. Thus, the number of selected assets for data gathering can be much higher than for single asset trading. On another note, we can observe the importance of the stock indexes as they being adopted directly as assets \cite{Yu2019,Pendharkar2018,Almahdi2017, Jeong2019, DengYue2017, Eilers2014} or indirectly for selection of stocks \cite{Maringer2014, ZhangJin2014, Zhang2016, JiangZhengyao2017, DingYi2018}.

\noindent\begin{minipage}{\textwidth}
{\small
\setlength{\LTleft}{-20cm plus -1fill}
\setlength{\LTright}{\LTleft}
\begin{longtable}{@{} p{0.28\textwidth} p{0.07\textwidth} p{0.1\textwidth} p{0.47\textwidth}  @{} }
\caption{Comparison of input data according, type of asset, the origin country of each asset, and a brief description of relevant information for the portfolio optimization problem.}\label{tab:data_2}\\

\toprule
\textbf{Single Asset Trading }\\
\toprule
\textbf{Reference} & \textbf{Type} & \textbf{Data From} & \textbf{Description} \\
\toprule

\cite{Zarkias2019} & FOREX & Europe, U.S. & EUR/USD currency pair \\ \hline
\cite{Ponomarev2019} & Index & Russia & RTS Index at the Moscow Exchange \\ \hline
\cite{Jeong2019} & Index & U.S., China, Europe, Korea & Indexes: S\&P 500 index (SP500), Hang Seng index (HSI), EuroStoxx50 index and Korea Stock Price index (KOSPI) \\ \hline
\cite{Wu2019} & Stocks & China & 6 stocks not specified \\ \hline
\cite{LiYang2019} & Stocks, Futures & China, U.S. & \begin{tabular}[t]{@{}l@{}} - Stocks (U.S.): Apple, IBM and P\&G \\ - Index-based futures: ES (U.S.), IF (China) \end{tabular} \\ \hline
\cite{GiacomazziDantas2018} & Stocks & Brazil & 22 Bovespa Stock Exchange stocks with largest market cap \\ \hline
\cite{Alimoradi2018} & Stocks & Iran & 20 Tehran Stock Exchange stocks, with high market cap and liquidity, from different sectors \\ \hline
\cite{Carapuco2018} & FOREX & Europe, U.S. & EUR/USD currency pair \\ \hline
\cite{Spooner2018} & Stocks & Not Specified & 10 stocks of varied market venues and sectors \\ \hline
\cite{DengYue2017} & Futures, Indexes & China, U.S., U.K., Hong Kong, Japan & Main: 3 Futures Contracts (China): index-based Stock-IF, commodities-based for silver and sugar. \newline
Global Mark. Exper.: S\&P 500 (U.S.), FTSE100 (U.K.),  Nikkei 225 (Japan), and others. \\ \hline
\cite{Si2017} & Futures & China & Stock-IF, Stock-IH and Stock-IC futures contracts \\ \hline
\cite{Feuerriegel2016} & Stocks & Germany & 331 stocks from CDAX and its text financial Disclosures \\ \hline
\cite{Gabrielsson2015} & Futures & U.S. & E-mini S\&P 500 index futures contracts \\ \hline
\cite{Eilers2014} & Indexes & Germany, U.S. & DAX (Germany) and S\&P 500 (U.S.) \\ \hline
\cite{ZhangJin2014} & Stocks & U.S. & 18 stocks from S\&P 500 \\ \hline
\cite{Maringer2014} & Stocks & U.S. & 238 stocks S\&P 500 without incomplete data \\
\bottomrule
\end{longtable}
}
\end{minipage}

\noindent\begin{minipage}{\textwidth}
{\small
\setlength{\LTleft}{-20cm plus -1fill}
\setlength{\LTright}{\LTleft}
\begin{longtable}{@{} p{0.23\textwidth} p{0.16\textwidth} p{0.31\textwidth}@{} }
\caption{Data input comparison according to the time-frame and period for the portfolio optimization problem.}
\label{tab:data_3}\\
\toprule
\textbf{Portfolio optimization}\\
\toprule
\textbf{Reference} & \textbf{Time-frame} & \textbf{Period (Approx. Data Points)} \\ \hline

\cite{Weng2020} & Intraday (30 min) & 2014 - 2017 (70080)  \\ \hline
\cite{Aboussalah2020} & Intraday (hour) & 2013 - 2017 (7928)  \\ \hline
\cite{Ye2020} & Stock: Daily \newline FOREX: Intraday (30 min.) & \begin{tabular}[p]{@{}l@{}}Stock.: 2006 - 2013 (35089)\\ Korea: 2015 - 2017 (1784) \end{tabular} \\ \hline
\cite{Park2019} & Daily & \begin{tabular}[p]{@{}l@{}}U.S.: 2010 - 2017 (1769)\\ Korea: 2012 - 2017 (1267) \end{tabular} \\ \hline
\cite{Almahdi2019} & Weekly & 2011 - 2015 (260) \\ \hline
\cite{Wang2019a} & Monthly & \begin{tabular}[p]{@{}l@{}}U.S.: 1970 - 2016 (564) \\China: Jun. 2005 - 2018 (163) \end{tabular} \\ \hline
\cite{Yu2019} & Daily & 2005 - 4 Dec. 2018 (3,476) \\ \hline
\cite{Pendharkar2018} & Quarterly, Semi-annual or Annual & S\&P 500, AGG: 1976 - 2016 (41); T-note: 1970 - 2016 (47, 77 and 137) \\ \hline
\cite{Kang2018} & Daily & 2010 - 31 Jul. 2017 (1,900) \\ \hline
\cite{DingYi2018} & Daily & 2005 - 2016 (2,946) \\ \hline
\cite{Almahdi2017} & Weekly & 2011 - 2015 (260) \\ \hline
\cite{JiangZhengyao2017} & Intraday (30 min.) & 27 Jun. 2015 - 27 Aug. 2016 (Est. 17,000) \\ \hline
\cite{Zhang2016} & Daily & 2009 - Apr. 2014 (1,340) \\
\bottomrule
\end{longtable}}
\end{minipage}

Moving on to Table \ref{tab:data_2}, we provide data characteristics regarding the time-frame of the collected data. One noteworthy aspect of the analyzed studies’ preferences of period and frequency is the interdependence between them. This dependency can occur because a lower frequency of data sampling may restrict the minimum number of data points or instances necessary for training a machine learning system that generalizes well. For this reason, we can observe a trend of gathering data across decades, with some coinciding periods.

\afterpage{\clearpage
{\small
\setlength{\LTleft}{-20cm plus -1fill}
\setlength{\LTright}{\LTleft}
\begin{longtable}{@{} p{0.23\textwidth} p{0.16\textwidth} p{0.31\textwidth}@{} }
\caption{Data input comparison according to the time-frame and period for the single asset problem.}
\label{tab:data_4}\\
\toprule
\textbf{Single Asset Trading}\\
\toprule
\textbf{Reference} & \textbf{Time-frame} & \textbf{Period (Approx. Data Points)} \\ \hline

\cite{Zarkias2019} & Intraday (4 Hour) & 2007 - 2015 (14000) \\ \hline
\cite{Ponomarev2019} & Intraday (Minute) & 2015 - 2016 (50000) \\ \hline
\cite{Jeong2019} & Daily & \begin{tabular}[p]{@{}l@{}}SP500: 1987 - 2017 (7796) \\HSI: 2005 - 2017 (4282) \\ EuroStoxx50: 1991 - 2017 (6731)\\ KOSPI: 1997 - 2017 (5163)\end{tabular}  \\ \hline
\cite{Wu2019} & Daily & 25 Oct. 2005 - 25 Oct. 2017 (2,954) \\ \hline
\cite{LiYang2019} & Intraday (Minute) & 2008 - 2017 (598,560) \\ \hline
\cite{GiacomazziDantas2018} & Daily & 2011 - 2017 (1,741) \\ \hline
\cite{Alimoradi2018} & Daily & Up to 20 Aug. 2016 (1,000) \\ \hline
\cite{Carapuco2018} & Intraday (Hour) & 2009 - 2017 (50 million ticks) \\ \hline
\cite{Spooner2018} & Daily & Jan - Aug. 2010 (160) \\ \hline
\cite{DengYue2017} & Intraday (Minute), Daily & Main experiments: \newline 2014 - Sep. 2015 (99,840) \newline Global Market Experiments: \newline 1990 - Sep. 2015 (6,500 days) \\ \hline
\cite{Si2017} & Intraday (Minute) & 2016 - Jun. 2017 (90,720) \\ \hline
\cite{Feuerriegel2016} & Daily & 2004 - Jul. 2011 (1956) \\ \hline
\cite{Gabrielsson2015} & Intraday (Minute) & 05 Jul. 2011 - 02 Sep. 2011 (58,800) \\ \hline
\cite{Eilers2014} & Daily & 2000 - 2012 (3251) \\ \hline
\cite{ZhangJin2014} & Daily & 2009 - 3 Dec. 2012 (980) \\ \hline
\cite{Maringer2014} & Daily & 2009 - 3 Dec. 2012 (980) \\ 
\bottomrule
\end{longtable}}
}

Some authors opted for lower frequency operations such as Daily \cite{Maringer2014, ZhangJin2014, Eilers2014, Zhang2016, DengYue2017, Alimoradi2018, DingYi2018, GiacomazziDantas2018, Kang2018,  Spooner2018, Wu2019, Yu2019}, Weekly \cite{Almahdi2017, Almahdi2019}, Monthly \cite{Wang2019a}, or higher than Quarterly \cite{Pendharkar2018}. Among these low frequencies works, even the highest one -- the Daily -- requires as much as ten years to reach about a thousand data points for each asset. Oppositely, papers that opted for intraday higher frequency data sampling (e.g., minutely, hourly, and others) can gather thousands of data instances within just one year period \cite{Gabrielsson2015, Si2017, DengYue2017, LiYang2019} and have less coinciding periods. Notwithstanding, one of the selected articles \cite{Pendharkar2018} compared using data on quarterly, semi-annual, and annual frequencies and, sometimes, observed better results on the latter one. For this reason, \cite{Pendharkar2018} concluded that more extensive datasets might lead to better results.

In general, we observe that among most works surveyed here, there is little uniformity over most of the data attributes we analyzed (e.g., assets' type, country, time-frame, period, and amount). This lack of standardization is meaningful to our analysis, and we will retake this aspect in Sections \ref{subsec:results} and \ref{subsec:general_standards}.

\subsection{Data preprocessing and state composition}
\label{subsec:preprocessing}

After selecting the time-series of prices of a particular asset, or a set of assets, it is usually necessary to make some treatment on the data extracted, as shown in Fig. \ref{fig:pipeline}, as a data preprocessing step. In Table \ref{tab:preprocessing_1} and Table \ref{tab:preprocessing_2}, we group preprocessing techniques into technical indicators, forecasting, standardization, encoding, and data cleaning. Then, we examine in Tables \ref{tab:state_1}, and \ref{tab:state_2} how the preprocessed features compose the state representation and look at the number of inputs for each type of feature and the total state size. Given the concepts discussed in Section \ref{subsec:contextual_bandit} we grouped articles that used a contextual bandit problem formulation in Tables \ref{tab:preprocessing_1} and \ref{tab:preprocessing_2}. Alternatively, for Tables \ref{tab:state_1} and \ref{tab:state_2}, we placed approaches that used full reinforcement learning.

\begin{figure}[ht]
    \centering
    \includegraphics[width=0.9\linewidth]{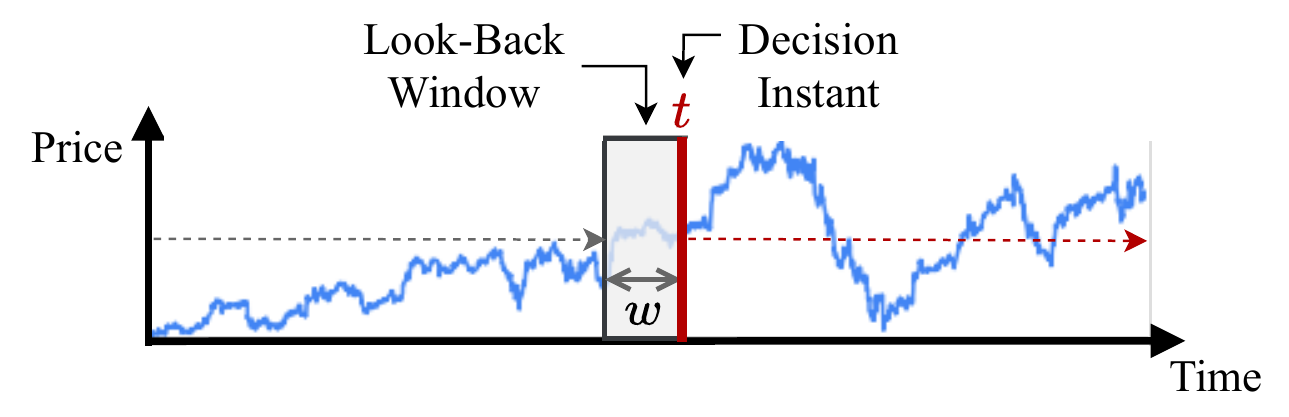}
    \caption{Price time-series with the sliding look-back window representation before the instant $t$.}
    \label{fig:look-back-window}
\end{figure}

For works that rely on time-series data, it is common to determine time-windows, which are sections of consecutive history data points. One particularly relevant window for our review is the look-back window (also known as the input or trading window). As shown in Fig. \ref{fig:look-back-window}, the \textit{look-back window} is a time-window of the past \(w\) price points used by the model, starting from a decision instant \(t\). Moreover, the look-back window moves through the price time-series as the model sequentially scans the data.

Initially, we found a crucial difference between most surveyed works and the RL approaches on the web marketing domain \cite{LihongLi2010, Theocharous2015} mentioned in Section \ref{subsec:contextual_bandit}. These marketing papers aimed at either a contextual bandit or a full RL formulation and only then designed their architectures accordingly. However, except for a minority of articles \cite{Eilers2014, Pendharkar2018, DingYi2018}, most of the surveyed works do not mention if they aimed at a specific formulation. Thus, to establish a classification criterion that maps how a researcher designs its state representation to the RL theoretical formulations described in Section \ref{subsec:contextual_bandit}, we draw inspiration from  \cite{Eilers2014, DingYi2018, Pendharkar2018}.

\cite{Pendharkar2018} assumed that, for a single investor trading modest amounts of assets, the impact of the actions on a large market is insignificant \cite{Hildenbrand1988}. With this simplification, \cite{Pendharkar2018} found that the problem at hand would be equivalent to a contextual bandit problem, in which the actions of the trading agent have no impact on the state. Consequently, \cite{Pendharkar2018} represented the complete system's states using exclusively market-state features. Two papers followed a similar assumption, \cite{Eilers2014, DingYi2018}, and adopted the same state representation. Therefore, inspired by the rationale of these three approaches, we establish that the absence of agent-state features is sufficient for placing a work into the contextual bandit group. Following this criterion, we considered full RL, authors that included any feature related to the agent condition such as the amount of assets \cite{Carapuco2018, Spooner2018}, or agent's previous actions \cite{DengYue2017, Almahdi2017, Almahdi2019}.

\cite{Wu2019} experimented with various indicators and, thus, combined these indicators into four different groups for better analysis. The group with the best results was a combination that included the Stochastics Oscillator with J and D line of calculation (KDJ), Relative Strength Index (RSI), and Williams Overbought/Oversold Index indicators. \cite{Carapuco2018} explored a different type of technical indicator, which considered the bid and ask prices with their respective volumes. Then, \cite{Carapuco2018} generated the mean, max, min, and standard deviation for all the technical indicators and later performed a further selection of preprocessed features to compose the state.

\textbf{Technical indicators:} We consider the generation of technical indicators as a preprocessing task. For instance, encoding is a preprocessing case that provides a dimensionality reduction of the data. Amid reviewed papers, at least half use technical indicators that compact the price time-series to a more concise state. We enumerate the preferred technical indicator categories among works as follows:

\begin{enumerate}

    \item Momentum Indicators: These methods' objective is to capture the rate-of-change of the series. Also, we observe that the Relative Strength Index (RSI) is a prevailing choice of amid articles \cite{Eilers2014,Zhang2016,Spooner2018,Alimoradi2018,Wu2019}. Some researchers \cite{Alimoradi2018,GiacomazziDantas2018} preferred the Stochastic Oscillator (SO) \cite{LoAndrew2000} as an alternative momentum indicator.
    \item Trend Indicators: These methods aim to identify the time-series' direction or tendency. The predominant indicator in this category is the Mean Average (MA), and its variants, such as the Moving Average Convergence/Divergence Oscillator (MACD) and Exponential Moving Average (EMA) \cite{Eilers2014,Alimoradi2018,GiacomazziDantas2018,LiYang2019,Wu2019}.
    \item Volume Indicators: Methods that help investors attain information about the trading volume (number of contracts or money traded). It is important to log at volume because buying and selling operations cannot occur without enough trading volume in the market. Techniques such as the positive indicator volume (IPVI), Indicator Negative Volume (INVI), Williams Overbought/Oversold Index, Volatility Volume Ratio (VVR), and Volume Ratio (VR) \cite{Zhang2016,Wu2019} can give information about the market volume trends. The volume indicator is the less used group of indicators since researchers usually assume high enough volume, implying high liquidity.
    \item Volatility Indicators: Methods that track volatility (also interpreted as price variance). They serve as guides to determine the possibly optimal exit or entry points for trades. In order to identify the market's volatility, some articles \cite{GiacomazziDantas2018,Wu2019} adopted indicators such as the Bollinger Bands \cite{bollinger2002bollinger}, while others \cite{LiYang2019} preferred the Average True Range (ATR).
    
\end{enumerate}

\textbf{Standardization (i.e., Normalization)}: It is a usual preprocessing step in machine learning in general. The vast majority of authors adopted the simple price difference between the sequence of data points as a standard practice. Alternatively, \cite{Carapuco2018} employed a percentile-based filter and normalized the return of price series in the range $[-1,1]$. Normalization is a common practice in time-series analysis and is also made clear by some authors \cite{JiangZhengyao2017,Almahdi2017,Kang2018,Carapuco2018,Aboussalah2020}, but its effectiveness in improving the learning model authors do not discuss.

\textbf{Encoding:} This is the transformation of data values to a schematic format that is easier to represent. Some works employed encoding also because of the dimensionality reduction that some of these techniques might produce. For instance, to reduce the inherent uncertainty and variability of price data, \cite{DengYue2017} employed fuzzy logic to convert the return series to a fuzzified representation. Another approach to encoding the input variables is the linear combination of tile codings (LCTC). \cite{Spooner2018} used LCTC to exploit variables' (approximate) independence. Notably, there is a recent tendency \cite{LiYang2019,Park2019,Ye2020} to adopt encoding techniques that employ artificial neural networks, called autoencoders.

\textbf{Data cleaning:} This preprocessing group of techniques is standard for fixing incorrect, inaccurate, or irrelevant data by replacing or modifying it. For example, \cite{JiangZhengyao2017} filtered the empty history data on cryptocurrency price time-series. In this particular case \cite{JiangZhengyao2017}, a method to generate a price matrix caused missing data to appear when cryptocurrencies did not exist. Therefore, before creating some of the cryptocurrencies, \cite{JiangZhengyao2017} filled the price time-series with a placeholder value and, thus, assumed there was no price change. In other cases \cite{LiYang2019}, data cleaning involved manually removing parts of data. During the initial open hours of market exchanges, events that occurred during closing hours might cause an abnormal fluctuation in the asset prices of affected companies. Thus, by removing the price time-series of some periods, \cite{LiYang2019} reduced the impact of these events.

\textbf{Forecasting: } We consider the price forecasting as a preprocessing step, given that it transforms the price time-series into a future price estimation that can compose the state representation. For instance, \cite{Yu2019} implements the Infused Prediction Module (IPM) to provide foresight of future prices. The idea is that from a more informative representation given by time-series forecasting, the agent can better approximate policies \cite{Yu2019}. Also, \cite{Yu2019} tested two different prediction models: a nonlinear dynamic Boltzman machine \cite{Dasgupta2017} and dilated convolution layers. The former is fast because it does not require backpropagation, and the WaveNet architecture \ inspires the latter cite{Oh2015}.

Finally, we identified some less adopted preprocessing methods and placed them in the group ``Other'', as seen in Tables \ref{tab:preprocessing_1} and \ref{tab:preprocessing_2}. For instance, among this group, we assumed preprocessing methods such as data augmentation \cite{Yu2019}, information acquisition by the behavior clone model \cite{Yu2019}, portfolio weights optimization using particle swarm optimization \cite{Almahdi2019}, bid aggregation \cite{Ponomarev2019}, generation of transition variables for a regime switch RRL \cite{Maringer2014}, and others.

{\small
\setlength{\LTleft}{-20cm plus -1fill}
\setlength{\LTright}{\LTleft}
\begin{longtable}{p{4cm}p{3,3cm}p{8cm}}
\caption{Articles with their respective preprocessing group and details/remarks of each work  for the contextual bandit problem.}
\label{tab:preprocessing_1}\\
\toprule
\textbf{Contextual Bandit}\\
\toprule
\textbf{Reference} & \textbf{Preprocessing Group} & \textbf{Details/Remarks} \\
\midrule
\cite{Weng2020} & Others & Uses XGboost to select best features \\
\hline
\cite{Zarkias2019} & Others & Uses a factor $\alpha$ that controls the internal estimation of the agent to update the state ant $t+1$  \\
\hline
\cite{Jeong2019} & No & No  \\
\hline
\cite{Wang2019a} & Tech. and Fund. Indicators; Encoding & Tech: Price, others; Fund.: PE, MC, BM, others; Encoding: Deep learning attention mechanism \\
\hline
\cite{Yu2019} & Forecasting; Others & IPM; DAM; BCM \\
\hline
\cite{GiacomazziDantas2018} & Technical Indicators & Return; MA; SO; Bollinger Bands; MACD \\
\hline
\cite{Alimoradi2018} & Technical Indicators & Golden/Dead Cross; Rate of Deviation; RSI; Volume Ratio; MACD; Ranked Correlation Index; Candlestick; Rate of Change; SO \\
\hline
\cite{Pendharkar2018} & No & No \\
\hline
\cite{DingYi2018} & Technical Indicators & Annualized Return; Max Drawdown; SR; ANN for feature extraction. \\
\hline
\cite{JiangZhengyao2017} & Standardization; Data Cleaning & 12 different cryptocurrency chosen buy volume-ranking \\
\hline
\cite{Feuerriegel2016} & Technical Indicators; Others & Rate-of-change (ROC), Sentiment Analysis of financial disclosures \\
\hline
\cite{Eilers2014} & Standardization; Tech. and Fund. Indicators & Tech.: RSI, MA, open, high, low, close prices, others; Fund.: turn-of-the-month, pre-FOMC announcement, others\\ \bottomrule 
\end{longtable}}

On another tone, we notice that although several researchers adopted continuous state spaces \cite{Pendharkar2018} adopted discrete states.
As observed in Tables \ref{tab:state_1} and \ref{tab:state_2}, the vast majority of the states receive the price value directly with only some simple prior preprocessing, such as standardization or data cleaning. Moreover, in some cases \cite{Zhang2016,JiangZhengyao2017,Carapuco2018,Jeong2019,Aboussalah2020,Weng2020}, the state size is equal to the look-back window size. This aspect is essential in determining policy approximation methods' state size and complexity. Some authors \cite{Eilers2014, Zhang2016, Wang2019a,Ye2020}, use a combination of technical and fundamental indicators to compose the state. \cite{Feuerriegel2016} employed \textit{sentiment analysis} (SA) techniques to extract features from textual news sources and use them in the state representation. Thus, this approach can be considered a fundamental analysis. \cite{Ye2020} also employed textual news but opted to adopt the Glove word embeddings \cite{Pennington2014} instead.

\noindent\begin{minipage}{\textwidth}
{\small
\setlength{\LTleft}{-20cm plus -1fill}
\setlength{\LTright}{\LTleft}
\begin{longtable}{p{4cm}p{9cm}p{1cm}}
\caption{List of articles with the respective state composition and state size for the contextual bandit problem.}
\label{tab:state_1}\\
\toprule
\textbf{Contextual Bandit}\\
\toprule
\textbf{Reference} & \textbf{Features (inputs) or States} & \textbf{State Size} \\
\midrule
\cite{Weng2020} & (3) Close, high and low prices with (n) look-back window & $3 \times n$ \\
\hline
\cite{Zarkias2019} & (1) Stock price & 1 \\
\hline
\cite{Jeong2019} & (200) Stock prices look-back window & 200 \\
\hline
\cite{Wang2019a} & (n) Winner score & n \\
\hline
\cite{Yu2019} & (8) Stocks - 10 time embedding & 80 \\
\hline
\cite{GiacomazziDantas2018} & (1) Last day Return; (1) Stochastic Oscillator; (1) MACD; (1) Bollinger Bands & 4 \\
\hline
\cite{Alimoradi2018} & Each state is a network graph of decisions based on previous technical indicators; Technical indicators consider mix of 5, 10 and/or 25 days & Not easy to estimate \\
\hline
\cite{Pendharkar2018} & (4) Discrete States E = {11, 01, 10, 00}, indicating which asset was negative or non-negative in the end of the evaluated period & 4 \\
\hline
\cite{DingYi2018} & Concatenation of 3 sets: smarket: feature extraction using tech indicators; smem: based on past experiences memory; spool: direct tech indicators of Exceed Return, AR (Annualized Return), Max Drawdown, SR, win rate & Not Specified \\
\hline
\cite{JiangZhengyao2017} & (12) number of assets x (50) number of time frames = (600) Total & 600 \\
\hline
\cite{Feuerriegel2016} & Current and previous disclosures: binary variables combination: (1) sign of sentiment; (1) rate-of-chance; (1) rate-of-change of index; (1) historic performance of index & 2 \\
\hline
\cite{Eilers2014} & (1) turn-of-the month or FOMC or exchange holidays; (4) open, high, low, close prices; (3) close price of past 3 days; (1) MA; (1) RSI; (1) Current month number &  \\ \bottomrule
\end{longtable}}
\end{minipage}

\noindent\begin{minipage}{\textwidth}
{\small
\setlength{\LTleft}{-20cm plus -1fill}
\setlength{\LTright}{\LTleft}
\begin{longtable}{p{4cm}p{3,3cm}p{8cm}}
\caption{Articles with their respective preprocessing group and details/remarks of each work for the full reinforcement learning problem.}
\label{tab:preprocessing_2}\\
\toprule
\textbf{Full reinforcement learning}\\
\toprule
\textbf{Reference} & \textbf{Preprocessing group} & \textbf{Details/Remarks} \\
\toprule
\cite{Aboussalah2020} & Standardization & Returns \\
\hline
\cite{Ye2020} & Encoding & Fundamental indicators from textual news using GloVe word embedding to compose the state \\
\hline
\cite{Park2019} & Encoding; Others & Uses long short-term memory autorencoder to compress sequence data; Generation of market features based on the OHCL, volume ratios and correlations. \\
\hline
\cite{Ponomarev2019} & Others & Aggregation of the bids \\
\hline
\cite{Almahdi2019} & Others & \textit{Many optimization liaisons} particle swarm optimization to select assets under constrains \\
\hline
\cite{Wu2019} & Technical Indicators & Different Techinical Indicators \\
\hline
\cite{LiYang2019} & Technical Indicators; Data Clearning & Technical indicators; Remaining trading cash Previous SR;  stacked denoising autoencoders; Remove the effect of news; Removed low volatility \\
\hline
\cite{Carapuco2018} & Standardization & Feature extraction of technical indicators considering time windows (TW) of candlesticks (Bid, Ask, Vol. of Bid and Vol of Ask) \\
\hline
\cite{Spooner2018} & Technical Indicators; Encoding & Linear combination of tile codings (LCTC), with: Market (bid-ask) spread (\(s\)), Mid-price move (\(\Delta m\)), Book/queue imbalance, Signed volume, Volatility, RSI \\
\hline
\cite{Kang2018} & Data Cleaning; Standardization & Nans replaces by zero; Min Max Scaling \\
\hline
\cite{DengYue2017} & Encoding & Fuzzy logic to encode price inputs \\
\hline
\cite{Almahdi2017} & Standardization & Log of the returns \\
\hline
\cite{Si2017} & Standardization; Others & 1) Uses multiple FC layers to capture features of the time-series \\
\hline
\cite{Zhang2016} & Standardization; Tech. and Fund. Indicators; Others & Tech.: IPVI, INVI, IRSI, others; Fund.: Others; Other: GA meta-heuristic for financial indicators selection \\
\hline
\cite{Gabrielsson2015} & Standardization; Technical Indicators & Seven different technical indicators \\
\hline
\cite{ZhangJin2014} & Standardization; Others & To determine the group for the multiple elitist, it uses the correlation between stocks \\
\hline
\cite{Maringer2014} & Standardization; Others & Calculate the transition parameters using GARCH and differential evolution for the Logistic-STAR model
\\
\bottomrule
\end{longtable}}
\end{minipage}

\noindent\begin{minipage}{\textwidth}
{\small
\setlength{\LTleft}{-20cm plus -1fill}
\setlength{\LTright}{\LTleft}
\begin{longtable}{p{4cm}p{9cm}p{1cm}}
\caption{List of articles with the respective state composition and state size for the full reinforcement learning problem.}
\label{tab:state_2}\\
\toprule
\textbf{Full reinforcement learning}\\
\toprule
\textbf{Reference} & \textbf{States Composition} & \textbf{State Size} \\
\toprule
\cite{Aboussalah2020} & 10 (number of stocks) $\times$ n (look-back window) & $10 \times n$ \\
\hline
\cite{Ye2020} & ($4 \times n$) open, high, low, close prices  \newline (100) News word embedding (1) predicted movement & $4\times n  + 101 $ \\
\hline
\cite{Park2019} & ($l \times n \times features(5)$) market features\newline where $n$ is the look-back window and $l$ is the number of assets & $l \times n \times features$ \\
\hline
\cite{Ponomarev2019} & Not specified & Not specified \\
\hline
\cite{Almahdi2019} & (3) log of price returns; (1) Previous action & 4 \\
\hline
\cite{Wu2019} & Different combinations of indicators sets: (10) d1, (31) d2, (34) d3, (35) d4; (1) Previous action & 10; 31; 34; 35 \\
\hline
\cite{LiYang2019} & (10) Technical indicators; (1) Previous SR; (1) Remaining trading cash & 12 \\
\hline
\cite{Carapuco2018} & (n) Look-back window from technical indicators; (2) Encoding of working hours; (1) current position; (1) Unrealized profit; (1) Size of the account compared to initial size. Author mentioned total size of 148 & 148 \\
\hline
\cite{Spooner2018} & (32) Linear combination of tile codings (LCTC) from Agent-State (Inventory, Act. Quoting Distances) and Market-State (Tech. Ind.) features & 32 \\
\hline
\cite{Kang2018} & (7500) Candlesticks (plus trading volume); (1) Previous action & 7501 \\
\hline
\cite{DengYue2017} & Return Values zt ($p_t - p_t-1$) look-back window: Main: 50 or 150; Global Market (S\&P500): 20; (1) Previous action & 51 or 151; 21 \\
\hline
\cite{Almahdi2017} & (105) log of price returns; (1) Previous action & 106 \\
\hline
\cite{Si2017} & (200) returns time-series; (1) Previous action & 201 \\
\hline
\cite{Zhang2016} & (7) Technical indicators; (10) Price returns; (1) Previous action & 18 \\
\hline
\cite{Gabrielsson2015} & (35) Technical Indicators; (1) Previous action & 36 \\
\hline
\cite{ZhangJin2014, Maringer2014} & (10) Stock prices look-back window; (1) Previous action & 11 \\
\bottomrule
\end{longtable}}
\end{minipage}

\subsection{Actions and reward functions}
\label{subsec:actions_rewards}

Following the workflow pipeline in Fig. \ref{fig:pipeline}, we now analyze the actions configuration and reward function of the surveyed articles. As described in Section \ref{sec:foundations}, given that the problem type (portfolio or single asset) influences the action space formulation, we accordingly divide works in Tables \ref{tab:actionreward_1} and \ref{tab:actionreward_2}. In these tables, we analyze action space aspects such as its type (i.e., continuous or discrete) and details about employed reward functions. For instance, some articles opted for continuous,  \cite{Maringer2014,Zhang2016,Almahdi2017,Ye2020,Aboussalah2020,Weng2020} while others preferred discrete \cite{Eilers2014,Gabrielsson2015} actions. 

For the portfolio optimization problem, the great majority of the works adopted discrete actions. The authors followed the premise that continuous actions are better suited for gradient methods, which can be policy approximation or actor-critic methods. On the other hand, all the RL methods can use discrete actions; therefore, some researchers also adopted discrete actions together with gradient methods.

\cite{Spooner2018} propose an alternative action space. In their work, the trading problem is the market-making problem. This problem aims to provide liquidity to the market by continuing to buy and sell securities without losing net worth. Given this slightly different problem formulation, their work has a different action space composed of a single buy or sell encoded vector with nine positions. The first eight positions represent pairs of buying or selling orders with a spread and bias relative to the prices. Then, the last actions are to clear inventory using a market order.

Analyzing the employed reward functions, we observe that nine reviewed works used the Sharpe ratio (SR) or the differential Sharpe ratio as a reward function \cite{Wang2019a,Pendharkar2018,DingYi2018,DengYue2017,Almahdi2017,Zhang2016,Gabrielsson2015,ZhangJin2014,Maringer2014}. 
SR is usually computed in a period (usually defined by the author for the specific problem) to measure profitability balanced by the risk. The use of differential SR as a reward function was first proposed by \cite{Moody1997} to adapt to online learning and speed the convergence properly. Another essential reward function is the return, expressed as the immediate return, logarithmic return, or profit. Some authors \cite{Eilers2014,JiangZhengyao2017,DengYue2017,Kang2018,Alimoradi2018,LiYang2019,Yu2019,Jeong2019,Ponomarev2019,Zarkias2019} calculated their returns based on the difference values in time. In order to calculate the return for the portfolio problem, some researchers \cite{JiangZhengyao2017,Kang2018,Pendharkar2018,Yu2019,Ye2020,Aboussalah2020,Weng2020} did as follows. The initial portfolio value compares with the final portfolio value. The final portfolio value is the portfolio value plus the total return in the period. The returns as a reward can be cumulative as it is in total (cumulative) profit. It can be an immediate reward, profit, or return obtained after the action resolution. Profit and reward are the most straightforward measures authors use.

Instead of avoiding volatility, several reward functions prefer evading the maximum drawdown or the downside deviation. For instance, \cite{Almahdi2019} uses the Calmar ratio, which is the relation between the return and the maximum drawdown. This measure tends not to focus on the price series volatility but concentrates on the maximum drawdown that would help the agent avoid excessive drawdown. Sortino ratio is another example, similar to the one used by \cite{Carapuco2018}, which, like the Sharpe ratio, is a relative risk metric that only considers the downside deviation. As \cite{Spooner2018} adopts a slightly different trading problem defined by the market-making the goal, their reward function presents a specific formulation. Profit and loss through the market exchanges depend on the volume of orders at a particular time and a mid-price. Moreover, their reward function also contains a component concerning the change in the agent’s cash holdings, given market price changes. Similarly, \cite{GiacomazziDantas2018} also created a customized reward function with another purpose, to amplify the impact of missed opportunities and risky situations avoidance.

\noindent\begin{minipage}{\textwidth}
{\small
\setlength{\LTleft}{-20cm plus -1fill}
\setlength{\LTright}{\LTleft}
\begin{longtable}{p{3.5cm}p{4.8cm}p{2cm}p{5.5cm}}
\caption{Works with the respective actions formulation, type of action (discrete or continuous), and type of reward for the portfolio optimization problem.}
\label{tab:actionreward_1}\\
\toprule
\textbf{Portfolio optimization} \\
\toprule
\textbf{Reference} & \textbf{Actions} & \textbf{Action Type} & \textbf{Reward Function} \\
\toprule
\cite{Weng2020} & Portfolio Vector (weights of the assets) & Continuous & The average logarithmic return of the Portfolio \\
\hline
\cite{Aboussalah2020} & Portfolio Vector (weights of the assets) & Continuous & The average logarithmic return of the Portfolio \\
\hline
\cite{Ye2020} & Portfolio Vector (weights of the assets) & Continuous & The average logarithmic return of the Portfolio \\
\hline
\cite{Park2019} & Long, neutral, short = {1,0,-1} & Discrete & Change in portfolio value (Immediate return) \\
\hline
\cite{Almahdi2019} & Long, short and neutral (A = [-1,1]) & Continuous & Calmar Ratio \\
\hline
\cite{Wang2019a} & Vector of binaries: {0,1} - invest or not invest in the stock & Discrete & Sharpe Ratio \\
\hline
\cite{Yu2019} & Portfolio Vector (weights of the assets) & Continuous & Immediate return: logarithmic rate of return \\
\hline
\cite{Pendharkar2018} & SARSA and Q Discrete Actions (5): - 0/100, 25/75, 50/50, 75/25, 100/0 ; TD Continuous Actions: - [0,1] proportion between assets & Both, depends on model & Sharpe Ratio, Total returns (used also for continuous actions) \\
\hline
\cite{Kang2018} & Branch 1: Portfolio Vector as actions; Branch 2: Value estimation & Continuous & Adjusted log reward of prices (Both branches) \\
\hline
\cite{DingYi2018} & Choice of logic M descriptor, which roughly translate to portfolio allocation & Continuous & Depends on the choice of type of investor: Oracle (Technical indicators); Collaborator (portfolio score); Public (revenue score) \\
\hline
\cite{Almahdi2017} & Portfolio Vector (weights of the assets) & Continuous & Differential Sharpe Ratio and Calmar Ratio \\
\hline
\cite{JiangZhengyao2017} & Portfolio Vector (weights of the assets) & Continuous & The average logarithmic return of the Portfolio \\
\hline
\cite{Zhang2016} & Long, short and neutral (A = [-1,1]) & Continuous & Differential Sharpe Ratio (DSR), for the indicators selection uses SR (Sharpe ratio))\\ \bottomrule

\end{longtable}}
\end{minipage}

Most reviewed methods optimize for one reward as the objective function. On the other hand, \cite{Si2017} proposes a multi-objective method that optimizes both the mean profit and the SR. Hence, with this method, it is also possible to use weight factors to balance goals.

\noindent\begin{minipage}{\textwidth}
{\small
\setlength{\LTleft}{-20cm plus -1fill}
\setlength{\LTright}{\LTleft}
\begin{longtable}{p{3.5cm}p{4.8cm}p{2cm}p{5.5cm}}
\caption{Works with the respective actions formulation, type of action (discrete or continuous), and type of reward for the single asset problem.}
\label{tab:actionreward_2}\\
\toprule

\textbf{Single Asset Trading} \\
\toprule
\textbf{Reference} & \textbf{Actions} & \textbf{Action Type} & \textbf{Reward Function} \\
\midrule
\endfirsthead
\cite{Zarkias2019} & Long, short and neutral (At = {1,0,-1}) & Discrete & Maintaining the control error equal to a arbitrary margin ($m$) \\
\hline
\cite{Ponomarev2019} & Long, short and neutral (At = {1,0,-1}) & Discrete & Immediate return \\
\hline
\cite{Jeong2019} & Buy, Sell, Hold depending on an threshold with number of share to be traded & Discrete & Immediate return \\
\hline
\cite{Wu2019} & Long, short and neutral (At = {1,0,-1}) & Discrete & Profit \\
\hline
\cite{LiYang2019} & positions-embedded action space: {-n,-n+1, ... 0, ... n-1, n} where n is maximum position & Discrete & Cumulative Profit \\
\hline
\cite{GiacomazziDantas2018} & long, short and neutral (At = {1,0,-1}) & Discrete & Proposed reward function: Formulation for rewarding agent for avoiding losses and penalizing for missing profit opportunities \\
\hline
\cite{Alimoradi2018} & Buy, Sell, Hold & Discrete & Reward $=$ selling price, buying price; Punishment $= 0.1 \times stock price$ (used for backward Q-learning in just specific conditions) \\
\hline
\cite{Carapuco2018} & Hold, Open or Close a Long or Short position: Only one unit of constant position size (10000 as experimentally defined) of chosen asset can be moved & Discrete & Normalized Sortino Based [-1,1], but with a min-max normalization to deal with weekends \\
\hline
\cite{Spooner2018} & 8 variations of pre-determined bid/ask amounts and 1 clear inventory & Discrete & Asymmetrically dampened profit and loss \\
\hline
\cite{DengYue2017} & Long, neutral, short = {1,0,-1} & Discrete & SR and total profit \\
\hline
\cite{Si2017} & Long, short and neutral (A = [-1,1]) & Continuous & Simultaneously both Average Profit and Sharpe Ratio. Weighted by factor. \\
\hline
\cite{Feuerriegel2016} & Long, short and neutral & Discrete & Cumulative profit \\
\hline
\cite{Gabrielsson2015} & Binary {+1,-1} long/short & Discrete & Differential Sharpe Ratio \\
\hline
\cite{Eilers2014} & Position: long, neutral, short; Holding Period: One day, two days; lever: one, two & Discrete & Profit of each order \\
\hline
\cite{ZhangJin2014} & Long, short and neutral (A = [-1,1]) & Continuous & Differential Sharpe Ratio \\
\hline
\cite{Maringer2014} & Long, short and neutral (A = [-1,1]) & Continuous & Differential Sharpe Ratio \\
\bottomrule
\end{longtable}
}
\end{minipage}

\subsection{Reinforcement learning methods}
\label{subsec:rl_methods}

\begin{figure}[ht]
\centering
\includegraphics[width=0.7\linewidth]{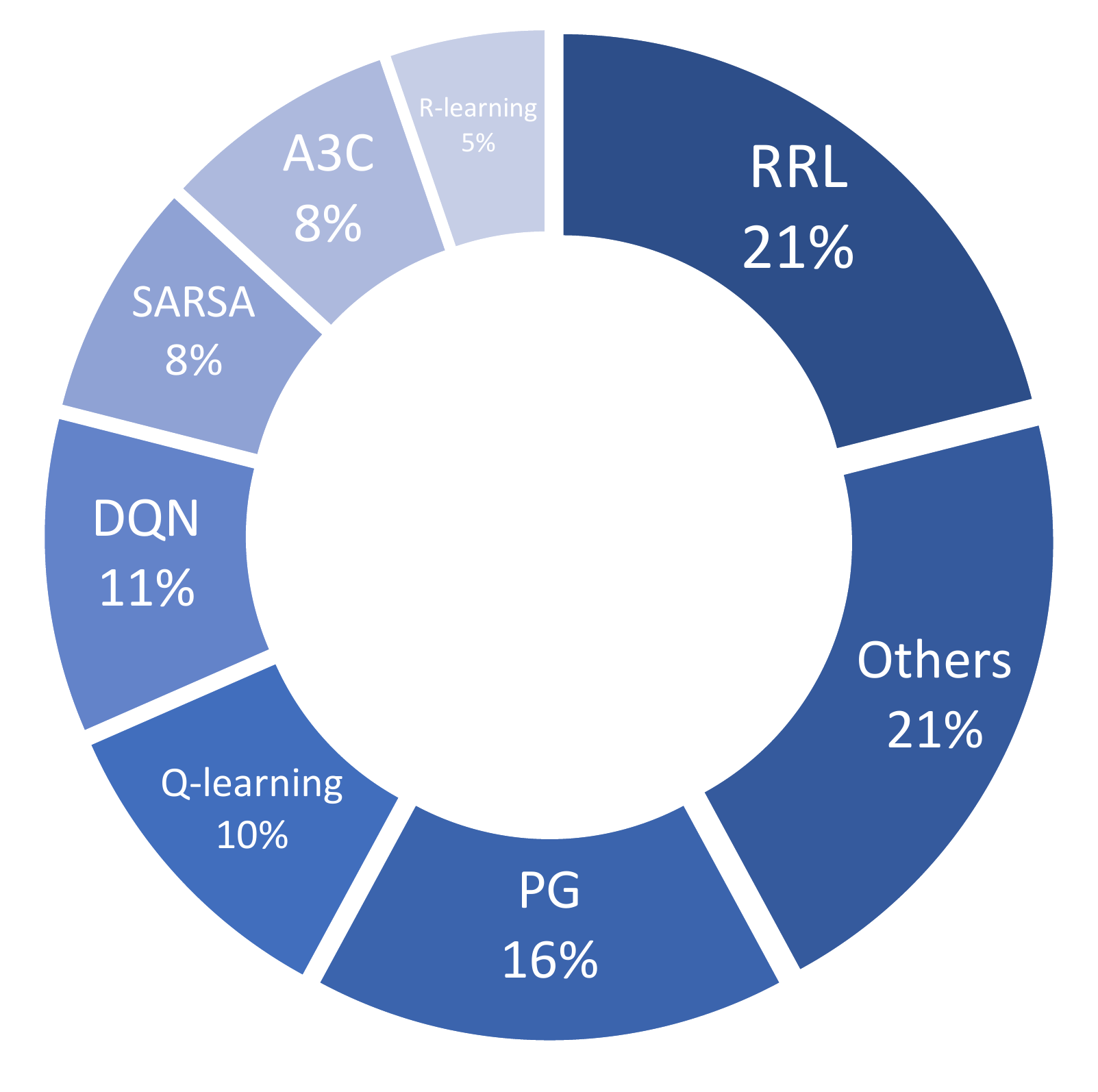}
\caption{Distribution of the RL methods over all the selected articles.}
\label{fig:methodsRLdistri}
\end{figure}

Recall from Section \ref{subsec:applied} that we presented a taxonomy for macroscopically dividing RL methods into the following categories: policy approximation (actor), TD-learning (critic), and hybrid techniques (or actor-critic). Also, we described how there are subgroups of techniques among these macroscopic categories, each with its particularities. Depicted in Fig. \ref{fig:methodsRLdistri} is the distribution of methods of the reviewed articles according to the mentioned subcategories. From this distribution of methods, it is notable that most reviewed papers adopt a policy approximation strategy, such as the policy gradient and recurrent reinforcement learning (RRL) method proposed by \cite{Moody1998}. We define RRL as a different policy gradient used as input from the previous actions and the return history. In the work of \cite{Moody1998}, the performance measure is the financial indicator Differential Sharpe Ratio formulated as:

\begin{equation}
S_t = \frac{E[R_t]}{\sqrt{E[R_t^2]-(E[R_t])^2}} = \frac{C}{\sqrt{B-C^2}},
\label{eq:diff_sharpe}
\end{equation}
and \(S_t\) here is analogous to the $J(\cdot)$ of the policy gradient in Eq. \ref{eq:01}, and $R_t$ is the return over the change of prices. 
Further, we show the partial derivatives of the Sharpe ratio, 
\begin{equation}
\frac{dS_t}{d\theta} = \frac{d}{d\theta} \Bigg\lbrace \frac{C}{\sqrt{B-C^2}} \Bigg\rbrace
\label{deriv_diff_sharpe_1}
\end{equation}
with
\[C=\frac{1}{T}\sum^T_{t=1}R_t\ \ \textrm{and}\ \ B=\frac{1}{T}\sum^T_{t=1}R_t^2\]
resulting in 
\begin{equation}
\frac{dS_t}{d\theta} = \sum^{T}_{t=1} \Bigg \lbrace \frac{dS_t}{dC}\frac{dC}{dR_t}+\frac{dS_t}{dB}\frac{dB}{dR_t}\Bigg\rbrace \cdot \Bigg\lbrace \frac{dR_t}{dA_t}\frac{dA_t}{d\theta}+\frac{dR_t}{dA_{t-1}}\frac{dA_{t-1}}{d\theta}\Bigg\rbrace,
\label{deriv_diff_sharpe_2}
\end{equation}
where the $\frac{dS_t}{d\theta}$ emerges as the updating factor of the model parameters, analogous to the gradient of the performance function $\nabla J(\theta)$.

Authors present some interesting variations of the RRL, such as in \cite{Maringer2014}. In their work, \cite{Maringer2014} employed regime switch variables (see also Section \ref{subsec:preprocessing}) to compose the update of the RRL parameters. \cite{Maringer2014} also adopted a differential evolution heuristic \cite{Storn1997} for optimization of transition parameters calculated for the Logistic-STAR model variation of the typical STAR model \cite{Taylor1998,Sarantis1999}. Moreover, in another work, \cite{ZhangJin2014} used a different update method for the RRL method. \cite{ZhangJin2014} considers a population of simulated traders with the initial parameters that undergo random initialization. Next, \cite{ZhangJin2014} generates the average elitist using the sampling of the highest Sharpe ratios and then calculates the mean parameters for the RRL parameters update. Besides, \cite{ZhangJin2014} proposes multiple elitist architectures for RRL parameters update, which considers the collective behavior of a group's correlated assets to adjust the RRL parameters.  

\cite{DengYue2017} refers to the RRL algorithm with the term direct reinforcement learning. Also, \cite{DengYue2017} explored the recurrent neural networks for policy approximation technique and fuzzy representation for feature learning. Interestingly, \cite{Moody2001} adopted the term direct reinforcement learning to refer to all algorithms that do not require value functions to derive the policy. Also called action-based algorithms, these approaches date back to \cite{Farley1954},  with extensions to these types of methods that include the REINFORCE \cite{Williams1992} and the direct gradient-based reinforcement \cite{Baxter1999}.

Other than adopting the RRL approach, some researchers alternatively apply the classical policy gradient as a policy approximation method \cite{JiangZhengyao2017,Deng2015,DingYi2018,Wu2019,Wang2019a,Aboussalah2020}. As an example, \cite{JiangZhengyao2017} investigates the combination of policy gradient with convolution neural networks in the portfolio optimization problem. \cite{Si2017} had a different approach with the Multi-Objective Direct Reinforcement Learning -- another terminology for a policy gradient method -- which adopted the daily profits standard deviation and mean, weighted by the factors $\alpha$ for the mean, and $\beta$ for standard deviation.

\noindent\begin{minipage}{\textwidth}
{\small
\setlength{\LTleft}{-20cm plus -1fill}
\setlength{\LTright}{\LTleft}
\begin{longtable}{p{3.5cm}p{7cm}p{5cm}p{4.5cm}}
\caption{Articles in the temporal difference category with their respective specific RL method adopted parameter update technique.}
\label{tab:taxonomy_1}\\
\toprule
\textbf{Temporal Difference}\\
\toprule
\textbf{Reference } & \textbf{RL Method} & \textbf{Parameter Update}\\
\midrule
\endfirsthead
\cite{Park2019} & Deep Q-network (DQN) & Not Specified \\
\hline
\cite{Zarkias2019} & Double  Deep Q-network (DDQN) & Adam \\
\hline
\cite{Jeong2019} & Deep Q-network (DQN) & Adam \\
\hline
\cite{GiacomazziDantas2018} & Q-Learning & Stochastic Gradient Descent \\
\hline
\cite{Alimoradi2018} & Backwards Q-learning (SARSA (on-policy) and Q-Learning (off-policy) hybrid) & Clasical bellman equation (uses Q values correspondent to tabular relations between nodes) \\
\hline
\cite{Pendharkar2018} & SARSA (on-policy), Q-Learning (off-policy) and TD learning (learns value functions directly, gradient descent algorithm) & - Q value function for SARSA and Q-learning \newline - Linear Gradient Descent for TD learning \\
\hline
\cite{Carapuco2018} & Q-Learning (author calls it Q-network instead of DQN) & Gradient Descent (No \(\varepsilon\)-greedy in testing) \\
\hline
\cite{Spooner2018} & Consolidated model: Expected SARSA (tested others, but this is the winner called consolidated model)- Double Q-learning- R-learning - Double R-learning - On-policy R-Learning  & Gradient Descent (exploration rate 0.7) \\
\hline
\cite{Feuerriegel2016} & Q-learning & Q-value function with \(\varepsilon\)-greedy of 0.2 \\
\hline
\cite{Eilers2014} & State-Action Pair Value function approximation without TD-learning & Gradient Descent \\ \bottomrule
\end{longtable}
}
\end{minipage}

The prevalent method amid surveyed works employed temporal difference learning \cite{Pendharkar2018,Carapuco2018,Spooner2018,Alimoradi2018,GiacomazziDantas2018,Jeong2019,Park2019,Zarkias2019} is Q-learning. Some articles \cite{Pendharkar2018,Spooner2018} adopted the SARSA method as an alternative to TD-learning. Interestingly, they also contrasted the performance of both methods. Among the diverse methods to approximate $Q$, some researchers preferred techniques such as neural networks \cite{Zarkias2019,Park2019,Jeong2019}, while others opted for different approximation methods \cite{Carapuco2018,Spooner2018,Alimoradi2018}.

\noindent\begin{minipage}{\textwidth}
{\small
\setlength{\LTleft}{-20cm plus -1fill}
\setlength{\LTright}{\LTleft}
\begin{longtable}{p{3.5cm}p{7cm}p{5cm}p{4.5cm}}
\caption{Articles in the policy approximation category with their respective specific RL method adopted parameter update technique.}
\label{tab:taxonomy_2}\\
\toprule
\textbf{Policy Approximation}\\
\toprule
\textbf{Reference} & \textbf{RL Method} & \textbf{Parameter Update} \\
\toprule
\cite{Aboussalah2020} & SDD RRL & Gradient Ascent\\
\hline
\cite{Almahdi2019} & RRL & Online Training Gradient Ascent \\
\hline
\cite{Wang2019a} & Policy Gradient & Gradient Ascent \\
\hline
\cite{Wu2019} & Policy Gradient & Adam (for the long short-term memory network) \\
\hline
\cite{DingYi2018} & REINFORCE & Gradient Descent (also uses e-greedy for selection of for selection of memory states during training) \\
\hline
\cite{DengYue2017} & Deep RRL & Backpropagation through time, online Training \\
\hline
\cite{Almahdi2017} & RRL & Gradient Ascent \\
\hline
\cite{Si2017} & Mult-objective deep reinforcement learning & Adam \\
\hline
\cite{JiangZhengyao2017} & Policy Gradient (CNN Agent) & Adam \\
\hline
\cite{Zhang2016} & RRL & Gradient Ascent \\
\hline
\cite{Gabrielsson2015} & RRL with lagged candlestick features & Gradient Ascent \\
\hline
\cite{ZhangJin2014} & RRL with Average Elitist and Multiple Elitist & Gradient Ascent, average elitist, multiple elitist \\
\hline
\cite{Maringer2014} & RS-RRL & Gradient Ascent, uses regime switch variables to update two parameters of the RRL method (as two models). Parameters are estimated with differential evolution and GARCH methods. \\ \bottomrule
\end{longtable}}
\end{minipage}

Next, we analyze works that employed a hybrid (i.e., actor-critic) approach that combines Temporal Difference (i.e., critic) and policy approximation (i.e., actor). In the works of \cite{Kang2018,LiYang2019,Ponomarev2019}, they adopted Asynchronous Advantage Actor-Critic (A3C) \cite{Mnih2016} as the RL method. \cite{Yu2019} uses a different actor-critic method called Deep Deterministic Policy Gradient (DDPG) \cite{Lillicrap2016}. In a deterministic policy gradient work, \cite{Silver2014} addressed the exploration issue for deterministic methods in a domain other than the financial one. \cite{Silver2014} argued that introducing noise can induce exploration for these methods, which is an aspect that \cite{Yu2019} explored. Nonetheless, \cite{Eilers2014} employed a deterministic policy approximation method and introduced exploration by forcing sporadic random actions. There are deterministic policy approximation works that do not disclose information about exploration mechanisms. Fortunately, to ensure exploration, the financial time-series used to simulate the market environment are noisy, which could be sufficient to expose the agent to diversified situations \cite{Silver2014}.

As explained in Section \ref{subsec:exploration_exploitation}, there are a variety of forms to address the trade-off between exploration and exploitation. Some TD articles adopted the \(\varepsilon\)-greedy mechanism to introduce some exploration in their RL methods \cite{Carapuco2018, Alimoradi2018, GiacomazziDantas2018, Spooner2018, Pendharkar2018}. Additionally, \cite{Carapuco2018} employed a varying \(\varepsilon\) parameter that decreased during training. Also, \cite{Alimoradi2018} introduced an exploration mechanism to the meta-heuristic part of its architecture by doing some artificial confrontations between better performing policies (exploitation) and lower performing policies (exploration). Other approaches used stochastic policy approximation methods that inherently handle exploration \cite{JiangZhengyao2017, Almahdi2017, Wang2019a, LiYang2019}. \cite{DingYi2018} adopted policy approximation and mentioned that using a preference memory -- that works similar to a replay memory -- would make even stochastic policies biased. For this reason, \cite{DingYi2018} employed the \(\varepsilon\)-greedy mechanism to increase exploration by selecting actions from the preference memory. \cite{LiYang2019} increased exploration by sampling random length episodes from the financial time series.

\noindent\begin{minipage}{\textwidth}
{\small
\setlength{\LTleft}{-20cm plus -1fill}
\setlength{\LTright}{\LTleft}
\begin{longtable}{p{3.5cm}p{7cm}p{5cm}p{4.5cm}}
\caption{Articles in the hybrid category with their respective specific RL method adopted parameter update technique.}
\label{tab:taxonomy_3}\\
\toprule
\textbf{Hybrid}\\
\toprule
\textbf{Reference} & \textbf{RL Method} & \textbf{Parameter Update} \\
\midrule
\cite{Ye2020} & DDPG & Stochastic Gradient Ascent \\
\hline
\cite{Ponomarev2019} & A3C & Gradient Descent \\
\hline
\cite{LiYang2019} & A3C and DQN & Gradient Descent \\
\hline
\cite{Yu2019} & Deep Deterministic Policy Gradient (DDPG) & Stochastic Gradient Ascent \\
\hline
\cite{Kang2018} & A3C and uses recurrent technique by taking last action & Gradient Descent \\ \bottomrule
\end{longtable}
}
\end{minipage}

\subsection{Setups for training and validation}
\label{subsec:setups}

\begin{figure}[ht]
    \centering
    \includegraphics[width=0.8\linewidth]{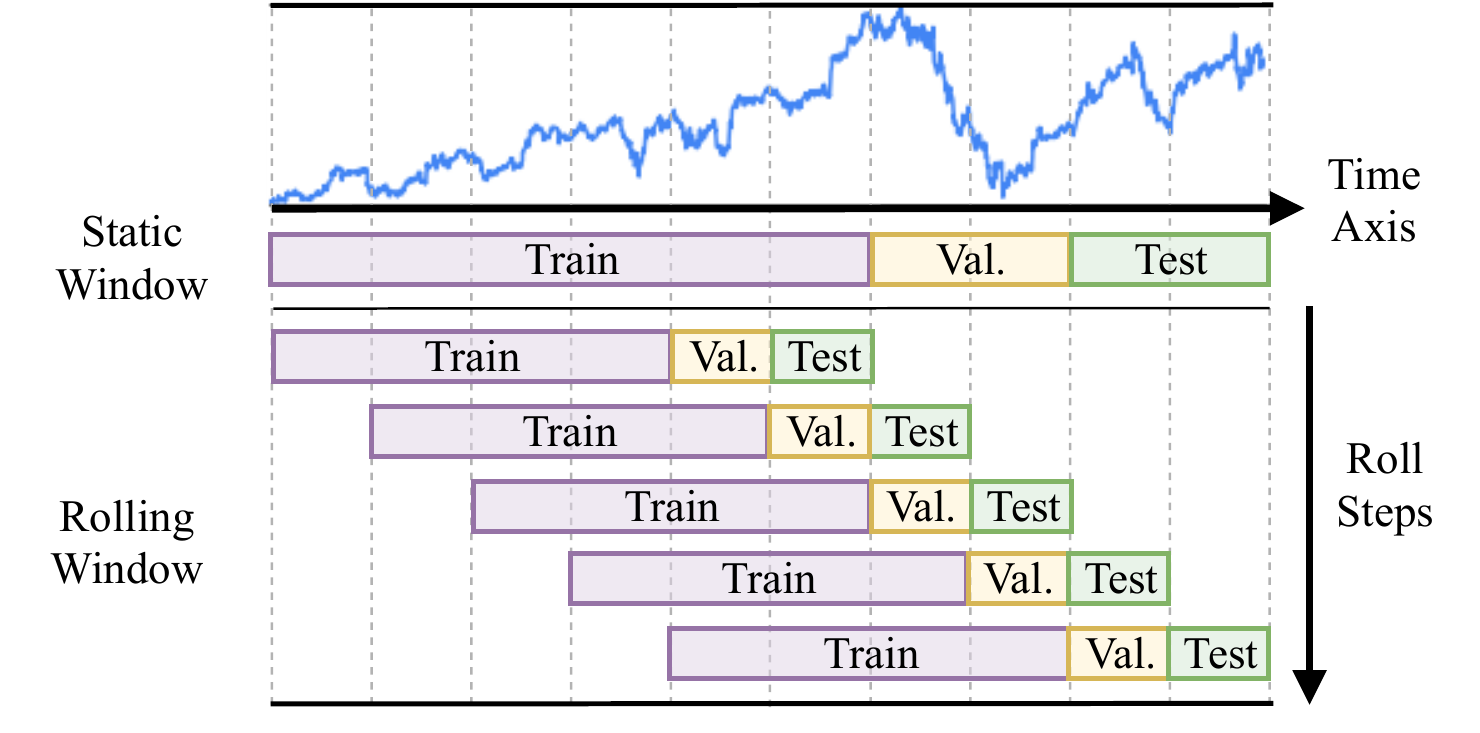}
    \caption{Comparison between static and rolling window configurations for the training, validation and test division.}
    \label{fig:rolling-window}
\end{figure}

In their study on overfitting in RL for video game tasks, \cite{ZhangChiyuan2018} discussed how it is usual for authors to execute continuous learning and evaluation under a single training scenario. However, \cite{ZhangChiyuan2018} argues that RL methods are not immune to overfitting and, thus, even more so for critical tasks, it is imperative to analyze the generalization capacity of an agent after training. For works that rely on time-series data, there are two general setups for evaluating the agent’s capacity to succeed in unseen conditions: the \textit{static window} and the \textit{rolling window}. In Fig. \ref{fig:rolling-window}, we have a comparison between these setups. Researchers partition their data samples in training, validation, and test sets for the static window (also known as a fixed window) setup. Subsequently, these sets correspond to specific, unchanging segments of the time-series. Oppositely, in the rolling window (also known as sliding or moving window \footnote{Financial researchers may be more familiar with the term walk forward \cite{Kirkpatrick2011} instead of rolling window.}) setup, there is a forward \textit{roll step} to guarantee training and evaluation of the model in different segments of data, containing different historical trends and aspects.

As observed in Fig. \ref{fig:rolling-window}, these setups exhibit different relevant characteristics and, thus, we separate in Table \ref{tab:static-window} the static window articles, while in Table \ref{tab:rolling-window}, we have rolling window ones. We note that one of the twenty-nine surveyed works did not specify any characteristics that allowed classification among the setup types and, thus, did not appear in either of the tables \cite{Eilers2014}. On the other hand, two of the reviewed articles employed both types of setup \cite{DengYue2017, Pendharkar2018}, and appear in both Tables \ref{tab:static-window} and \ref{tab:rolling-window}.

Among these two works, \cite{DengYue2017} initially applied the rolling window on the main experiments and, then used the static window on the global market experiment (recall that we explained the difference between these work's experiments in Section \ref{subsec:raw_data}). Alternatively, \cite{Pendharkar2018} employed the expansion window, a slight variation of the rolling window, where the training window expands at each window roll, and retraining of the machine learning model occurs at each roll before the evaluation step on a new test sample. Curiously, \cite{Pendharkar2018} proposed to use these types of setups as a part of the model functionality. In this sense, this work compared the so-called \textit{adaptive agent}, which used the expansion window setup, with the agent's static version. In this comparison, the adaptive agent performed better than its static counterpart but with the negative side of having a significantly higher computational time complexity.

The information present in Tables \ref{tab:static-window} and \ref{tab:rolling-window} helps identify the conditions authors subjected their agents. Thus, in Table \ref{tab:static-window}, there are details such as selected periods for training/validation/test sets and their distribution. Moreover, the period's choice for each set can influence the examination of various aspects, such as the used RL methods' generalization capacity. For instance, \cite{Wang2019a} tested their agent during both the U.S. \textit{dot-com bubble} (from 2000 to 2002) and the \textit{housing crisis} (2007 to 2009) events because they viewed these market instabilities as beneficial for evaluating the system adaptability to diverse situations. Similarly, \cite{DingYi2018} purposefully placed an unstable Chinese market period in its test set to examine the effects of the \textit{Chinese stock market turbulence} (June 2015 to February 2016). The mentioned episodes are examples of high-impact market events that can alter the financial market's behavior and cause unexpected changes in asset prices. Almost sixty percent of all twenty-nine reviewed works, trained or evaluated their models with data that included some portion of a crisis period \cite{Maringer2014, ZhangJin2014, Zhang2016, Feuerriegel2016, Zhang2016, Si2017, JiangZhengyao2017, Pendharkar2018, Carapuco2018, DingYi2018, Wang2019a, Wu2019, LiYang2019, Yu2019, Zarkias2019, Aboussalah2020, Ye2020}.

\noindent\begin{minipage}{\textwidth}
{\footnotesize
\setlength{\LTleft}{-15cm plus -1fill}
\setlength{\LTright}{\LTleft}
\begin{longtable}{@{} p{0.1\textwidth} p{0.25\textwidth} p{0.2\textwidth} p{0.2\textwidth} p{0.3\textwidth} @{} }

\caption{Static window setup characteristics according to training, validation, and test period windows and overall data distribution of sets. \textbf{Note:} We devised reasonably accurate estimates for articles that did not specify some of the information here.}
\label{tab:static-window}\\

\toprule
\multicolumn{5}{l}{Static Window}\\
\toprule
\textbf{Reference} &
\textbf{Train Period} &
\textbf{Validation Period} &
\textbf{Test Period} &
\textbf{Data Distribution (Ratio)} \\
\toprule
  
\cite{Ye2020} & 
  FOREX: 2015 - 2017 \newline Stock: 2006 - 2013 &
  No &
  FOREX: 2017 \newline Stock: 2013 &
  FOREX:(30177/-/4912) (0.86/-/0.14) \newline Stock:(1409/-/375) (0.79/-/0.21) \\ \hline
  
\cite{Zarkias2019} & 
  2007 - 2014 &
  No &
  2014 - 2015 &
  (12040/-/1960) 4 hours \newline
  (0.88/-/0.12) \\ \hline
  
\cite{Park2019} & 
  U.S: 2010 - 2016 \newline  Korean: 2012 - 2016 &
  No &
  U.S: 2017 \newline  Korean: 2017 &
  U.S: (1521/-/248) days \newline  Korean: (1090/-/177) days \newline(0.86/-/0.14) \\ \hline

\cite{Ponomarev2019} & 
  15 Set. 2015 - 15 Dec. 2015 &
  No &
  15 Dec. 2015 - 15 Jun. 2016 & (16500/-/33500) 30 min.
  (0.33/-/0.67) \\ \hline
  
\cite{Jeong2019} &
  SP500: 1987 - 2002  \newline
  HSI: 2001 - 2008\newline
  EuroStoxx50: 1991 - 2003 \newline
  KOSPI: 1997 - 2006&
  SP500: 2002 - 2006  \newline
  HSI: 2008 - 2009\newline
  EuroStoxx50: 2003 - 2005 \newline
  KOSPI: 2006 - 2008 &
  SP500: 2006 - 2017  \newline
  HSI: 2009 - 2017\newline
  EuroStoxx50: 2003 - 2017 \newline
  KOSPI: 2008 - 20017 &
  SP500: Est. (3742/936/3118) days \newline
  HSI: Est. (2055/514/1713) days\newline
  EuroStoxx50: Est. (3231/808/2692) days \newline
  KOSPI: Est. (2478/620/2065) days \newline
  (0.48/0.12/0.4) \\ \hline
  
\cite{Almahdi2019} &
  2011 - 14 Jan. 2014 &
  No &
  15 Jan. 2014 - 2015 &
  158/102 weeks (0.6/0.4) \\ \hline
\cite{Wang2019a} &
  U.S. Market: 1970 - 1989 \newline China: Jun. 2005 - 2011 &
  Values not specified &
  U.S. Market: 1990 - 2016 \newline China: 2012 - 2018 &
  U.S.: 240/-/324 (0.4/-/0.6) \newline China: 79/-/ 84 (0.5/-/0.5) \\ \hline
\cite{Yu2019} &
  2005 - 2016 (expanded to 30000 Synthetic data points) &
  No &
  2017 - 4 Dec. 2018 &
  Est.: 2996/480 days (0.9/0.1) \\ \hline
\cite{LiYang2019} &
  2008 - 2016 &
  No &
  2017 &
  538704/59856 min.(0.9/0.1) \\ \hline
\cite{GiacomazziDantas2018} &
  2011 - 2016 &
  No &
  2017 &
  Est.: 1493/248 days (0.85/0.15) \\ \hline
\cite{Alimoradi2018} &
  Estimate: Aug. 2012 - Aug. 2015 &
  No &
  Estimate: Aug. 2015 - 20 Aug. 2016 &
  750/250 days (0.75/0.25) \\ \hline
\cite{Pendharkar2018} &
  1976 - 2001 (S\&P 500, AGG) \newline 1970 - 1999 (T-note) &
  No &
  2002 - 2016 (S\&P 500, AGG) \newline 1970 - 2016 (T-note) &
  S\&P 500, AGG: 26/15 years (0.65/0.35) \newline T-note: 15/17 years, 60/34 sem. or 120/68 quarters (0.65/0.35) \\ \hline
\cite{Spooner2018} &
  Estimate: Jan - Jun 2010 &
  Values not specified &
  Estimate: Jul. - Aug. 2010 &
  120/-/40 days (0.75/-/0.25) \\ \hline
\cite{Kang2018} &
  2010 - 2013 &
  No &
  2014 - Jul. 2017 &
  1000/900 days (0.53/0.47) \\ \hline
\cite{DingYi2018} &
  2005 - 2013 &
  No &
  2014 - 2016 &
  Est.: 2202/744 days (0.75/0.25) \\ \hline
\cite{DengYue2017} &
  Global Exp.: 1990 - Oct. 1997 &
  Global Market Experiment: No &
  Global: Nov. 1997 - Sep. 2015 &
  Global Exp.: 2000/4500 days (0.3/0.7) \\ \hline
\cite{Almahdi2017} &
  2011 - 2013 &
  No &
  2014 - 2015 &
  156/104 weeks (0.6/0.4) \\ \hline
\cite{Si2017} &
  2016 - Mar. 2017 &
  No &
  Apr. - Jun. 2017 &
  Est.: 76080/14640 min. (0.85/0.15) \\ \hline
\cite{Feuerriegel2016} &
  2004 - 2005 &
  No &
  2006 - Jul. 2011 &
  Est.: 500/1456 days (0.25/0.75) \\ \hline
\cite{Zhang2016} &
  Estimate: 2009 - Oct. 2013 &
  Est.: May - Oct. 2013 &
  Est.: Nov. 2013 - Apr. 2014 &
  Est.: 1215/125/125 days (0.8/0.1/0.1) \\ \hline
\cite{ZhangJin2014, Maringer2014} &
  Est.: 2009 - 2010 &
  Est.: 2011 &
  Est.: Jan. - 3 Dec. 2012 &
  500/250/230 days (0.5/0.25/0.25) \\ \bottomrule
\end{longtable}}
\end{minipage}

Table \ref{tab:rolling-window} shows that about a quarter of reviewed articles experimented with a rolling window setup \cite{Gabrielsson2015, JiangZhengyao2017, DengYue2017, Carapuco2018, Pendharkar2018, Wu2019, Aboussalah2020, Weng2020}. Interestingly, \cite{Carapuco2018} advocates for using a rolling window setup by reasoning that because markets are non-stationary, differences in circumstances between sets may yield inaccurate results. Besides, as already mentioned, \cite{Pendharkar2018} executed a comparison between window setups that seemed to favor the use of the rolling window. Additionally, \cite{JiangZhengyao2017} verified the benefits of having the training set consistently close to the set being tested on, even suggesting that online training could be more suitable. In Section \ref{subsec:general_standards}, we discuss our views about all these aspects.

\noindent\begin{minipage}{\textwidth}
{\small
\setlength{\LTleft}{-20cm plus -1fill}
\setlength{\LTright}{\LTleft}
\begin{longtable}{@{} p{0.17\textwidth} p{0.275\textwidth} p{0.275\textwidth} p{0.17\textwidth} p{0.23\textwidth} @{} }
\caption{Rolling window setup characteristics according to training, validation and test periods of first/last windows and data distribution of sets during roll steps. \textbf{Note:} We devised reasonably accurate estimates for articles that did not specify some of the information here.}
\label{tab:rolling-window}\\

\toprule
\textbf{Rolling Window} \\
\toprule
\textbf{Reference} &
\textbf{First Window} &
\textbf{Last Window} &
\textbf{Data Distribution (Ratio)} &
\textbf{Roll Step Details} \\
\toprule
\cite{Weng2020} &
  Fist Validation:\newline  Tr.: Feb. 2014 - Dez. 2015 \newline  Test.: Dez. 2015 - Feb. 2016 &
  Last back test:\newline  Tr.: Feb. 2016 - Dez. 2017 \newline  Test.: Dez. 2017 - Feb. 2018 &
  Est. 32300/-/2800 half-hours (0.92/0.0.08) &
  Roll Step Size: 4 \newline Number of rolls: 6 \\ \hline
\cite{Aboussalah2020} &
  Tr.: Jan. 2005 - Jan. 2006 \newline  Test.: Oct. 2005 - Dec. 2006 &
  Tr.: Jan. 2016 - Jan. 2017 \newline  Test.: Oct. 2016 - Dec. 2017 &
  1200/-/300 hours (0.8/0.2) &
  Roll Step Size: 1 year \newline Number of rolls: 20 \\ \hline
\cite{Wu2019} &
  Tr.: 25 Oct. 2005 - 24 Oct. 2007 \newline Val.: 25 Oct. 2007 - 24 Oct. 2008 \newline Test: 25 Oct. 2009 - 24 Oct. 2010 &
  Tr.: 25 Oct. 2013 - 24 Oct. 2014 \newline Val.: 25 Oct. 2015 - 24 Oct. 2016 \newline Test: 25 Oct. 2016 - 24 Oct. 2017 &
  Est.: 438/-/219 days (0.66/-/0.33) &
  Roll Step Size: 1 year \newline Number of rolls: 9 \\ \hline
\cite{Pendharkar2018} &
  Train: 1976 - 2001 (S\&P 500, AGG), 1970 - 1999 (T-note) \newline Test: Next year, sem. or quarter &
  Train: 1976 - 2015 (S\&P 500, AGG), 1970 - 2015 (T-note) \newline Test: 2016 &
  Not applicable because train set expands when rolling and test set is only next data point. &
  Roll: 1 year, sem. or quarter \newline N. of rolls: 15-17, 30-34, 60-68 (depending on roll time unit) \\ \hline
\cite{Carapuco2018} &
  Train: 2009 \newline Val.: Jan. 2010 - Jun. 2010 \newline Test: Jul. 2010 - Oct. 2010 &
  Train: Mar. 2016 - Fev. 2017 \newline Val.: Mar. 2017 - Aug. 2017 \newline Test: Sep. 2017 - Dec. 2017 &
  12/6/4 months (0.7/0.2/0.1), with millions of data points each month &
  Roll Step Size: 4 months \newline Number of rolls: 23 \\ \hline
\cite{DengYue2017} &
  Main Experiment \newline Train: Jan - 19 Fev. 2014 \newline Val.: 20 Fev. - 5 Mar. 2014 \newline Test: 5 Mar. - 25 Mar. 2014 &
  Main Experiment \newline Train: 19 Jun. - 28 Jul. 2015 \newline Val.: 29 Jul. - 9 Aug. 2015 \newline Test: 10 Aug. - Aug. 2015 &
  12000/3000/5000 min. (0.6/0.15/0.25) &
  Roll Size: 5000 min. (20 days) \newline Number of rolls: 17 \newline Notes: Dates are our estimates. \\ \hline
\cite{JiangZhengyao2017} &
  Tr.: 27 Jun. 2015 - 14 Mar. 2016 \newline Val.: 3 May - 27 Jun. 2016 \newline Test: 14 Mar. - 3 May 2016 &
  Tr.: 27 Aug. 2015 - 14 May 2016 \newline Val.: 3 Jul. - 27 Aug. 2016 \newline Test: 14 May - 3 Jul. 2016 &
  12000/2500/2500 (0.75/0.15/0.15) half-hours grouping datapoints &
  Roll Step Size: 1 month \newline Number of rolls: 3 \\ \hline
\cite{Gabrielsson2015} &
  Train: 6 Jul. - 18 Jul. 2011 \newline Val.: 19 Jul. 2011 \newline Test: 20 Jul. 2011 &
  Train: 18 Aug. 2011 \newline Val.: 1 Sep. 2011 \newline Test: 2 Sep. 2011 &
  14000/1400/1400 min. (0.8/0.1/0.1) &
  Roll Size: 1,400 min. (1 day) \newline Number of rolls: 31 \newline Note: Dates are our estimates. \\
\bottomrule
\end{longtable}}
\end{minipage}

As the amount of data used for training can influence the generalization capacity of reinforcement learning models \cite{ZhangChiyuan2018}, we observe from both Tables \ref{tab:static-window} and \ref{tab:rolling-window} that most works adopted a data distribution that places most available data in the train set. Some papers discussed the validation set is essential to avoid model overfitting and for adequate selection of the model's hyperparameters, both of which could increase performance on the test set \cite{Gabrielsson2015, DengYue2017, Carapuco2018}. Nevertheless, only six static windows \cite{Maringer2014, ZhangJin2014, Zhang2016, Spooner2018, Jeong2019, Wang2019a}, and five rolling windows \cite{Gabrielsson2015, JiangZhengyao2017, DengYue2017, Carapuco2018, Wu2019} articles employed such practice. Still, this decision to avoid a validation set could be related to limited data availability, depending on the trading frequency, for composing an additional set.

Overall, the noticeable lack of similarity regarding input data characteristics observed in Section \ref{subsec:raw_data} is also present in the training and evaluation properties exhibited in Tables \ref{tab:static-window} and \ref{tab:rolling-window}. For instance, amid surveyed works, only two articles from the same researchers \cite{ZhangJin2014, Maringer2014}, share matching aspects such as train/test periods, data distribution, and adopting a validation set. This absence of standards can be crucial for adequately comparing results among studies and, thus, is a subject that we further examine in Section \ref{subsec:results}.

\subsection{Results, findings, and contributions of surveyed articles}
\label{subsec:results}

The reviewed articles’ common objective was to identify the capability of learning a trading strategy capable of profiting from the constant systematic fluctuations of the market (i.e., Arbitrage). In this regard, all authors reported success in the design of profitable systems. This overall profitability is noticeable according to the positive \textit{annualized return} (AR), and \textit{Sharpe Ratio} (SR) values present in Tables \ref{tab:results-metrics_1} and \ref{tab:results-metrics_2}. Among researchers that reported the AR, we have an average value of about 22.13\%, maximum and minimum values of 71.56\% \cite{Alimoradi2018}, and 5.71\% \cite{Kang2018}, respectively. Also, the average SR among authors that reported this risk-adjusted metric was 1.203. Some authors \cite{Spooner2018, Jeong2019, Zarkias2019} evaluated profitability using metrics (e.g., total return or absolute profit) that are not comparable amid different data periods. In Tables \ref{tab:results-metrics_1} and \ref{tab:results-metrics_2}, we list, but do not report, the values of these and other metrics.

\noindent\begin{minipage}{\textwidth}
{\small
\setlength{\LTleft}{-20cm plus -1fill}
\setlength{\LTright}{\LTleft}
\begin{longtable}{@{} p{0.28\textwidth} p{0.15\textwidth} p{0.15\textwidth} p{0.42\textwidth} @{} }
\caption{Results values for the single asset problem according to the average annualized return (AR), and the Sharpe ratio (SR). Also list of other adopted evaluation metrics. \textbf{Note:} For clarity and consistency of our work, we have estimated, for particular works, the AR or/and the SR with reasonable accuracy.}
\label{tab:results-metrics_2}\\
\toprule
\textbf{Single Asset} & & &\\
\toprule

\textbf{Reference} & \textbf{Avg. AR (\%)} & \textbf{SR} & \textbf{Other metrics} \\ \hline
\toprule
\cite{Zarkias2019} & 24.375 & - & Total Return \\ \hline
\cite{Ponomarev2019} & 29.535 & 1.774 & Fraction of winning transactions \\ \hline
\cite{Jeong2019} & SP500: 65.6, KOSPI: 56.7, HSI: 92.6, Eurostoxx50: 55.8 & - & PnL (Profit and Loss) \\ \hline
\cite{Wu2019} & 10.5 & - & Total Profit \\ \hline
\cite{LiYang2019} & Est. 40.586 & Est. 3.44 & Total Profit \\ \hline
\cite{GiacomazziDantas2018} & 14.21 & - & Max Drawdown, Exceed Return, Win Rate \\ \hline
\cite{Alimoradi2018} & 71.56 & - & Total Profit \\ \hline
\cite{Carapuco2018} & Est. 13.0 & - & Max Drawdown \\ \hline
\cite{Spooner2018} & - & - & Total Profit, Number of trades \\ \hline
\cite{DengYue2017} & - & 0.1643 & Portfolio Value, Max Drawdown \\ \hline
\cite{Si2017} & - & 0.1133 & Standard Deviation of SR \\ \hline
\cite{Feuerriegel2016} & - & 0.1168 & Cumulative Return, Volatility \\ \hline
\cite{Gabrielsson2015} & - & 0.00022 & - \\ \hline
\cite{Eilers2014} & SP500: 4.66, DAX: 8.04 & - & Max Drawdown \\ \hline
\cite{ZhangJin2014} & - & 0.0414 & Standard Deviation of SR \\ \hline
\cite{Maringer2014} & - & Est. 0.05 & Information Ratio, Standard Deviation of SR \\ \hline
\end{longtable}}
\end{minipage}

Recall that we pointed out before the lack of standards among articles regarding input data characteristics (Section \ref{subsec:raw_data}) and the training and evaluation of systems (Section \ref{subsec:setups}). This absence of uniformity reflects a variety of market scenarios, and the aforementioned positive results show broad market exploitability. On the other hand, to confidently recognize any specific exploitability of a market over a certain period, various articles would have to adopt standards (i.e., data systematically

\noindent\begin{minipage}{\textwidth}
{\small
\setlength{\LTleft}{-20cm plus -1fill}
\setlength{\LTright}{\LTleft}
\begin{longtable}{@{} p{0.28\textwidth} p{0.15\textwidth} p{0.15\textwidth} p{0.42\textwidth} @{} }
\caption{Results values for the portfolio optimization problem according to the average annualized return (AR), and the Sharpe ratio (SR). Also list of other adopted evaluation metrics. \textbf{Note:} For clarity and consistency of our work, we have estimated, for particular works, the AR or/and the SR with reasonable accuracy.}
\label{tab:results-metrics_1}\\
\toprule
\textbf{Portfolio optimization} & & & \\
\toprule
\textbf{Reference} & \textbf{Avg. AR (\%)} & \textbf{SR} & \textbf{Other metrics} \\ \hline
\endhead
\cite{Weng2020} & -  & 0.1 & Average return, Max Drawdown \\ \hline
\cite{Aboussalah2020} & Est. 19.58 & - & Total Profit \\ \hline
\cite{Ye2020} & - & FOREX: 13.89 \newline Stock: 3.83 & Portfolio Value \\ \hline
\cite{Park2019} & US: 12.634 \newline Korea: 9.933 & US: 2.071 \newline
Korea: 0.946 & Cumulative return, Sterling ratio, Average turnover \\ \hline
\cite{Almahdi2019} & 9.49 & 1.648 & Max Drawdown, Sterling Ratio \\ \hline
\cite{Wang2019a} & U.S.: 14.3; China: 12.5 & U.S.: 2.132; China: 1.22 & Annualized Volume, Max Drawdown, Calmar Ratio, Downside Deviation Ratio \\ \hline
\cite{Yu2019} & 8.09 & 0.63 & Annualized Volume, Sortino Ratio, Max Drawdown, Value-at-risk (VaR), Conditional value-at-risk (CVaR) \\ \hline
\cite{Pendharkar2018} & Est. 17.5 & - & Total Profit \\ \hline
\cite{Kang2018} & Est. 5.71 & - & - \\ \hline
\cite{DingYi2018} & 0.404 & 0.0883 & - \\ \hline
\cite{Almahdi2017} & 13.4 & 1.93 & Total Profit \\ \hline
\cite{JiangZhengyao2017} & - & 0.037 & Portfolio Value \\ \hline
\cite{Zhang2016} & - & Est. 0.05 & Profit, Number of profitable trades, Max Drawdown \\ \hline
\end{longtable}}
\end{minipage}

Concerning market efficiency, this absence of standards forbids us from confidently asserting markets' exploitability by directly comparing results in Tables \ref{tab:results-metrics_2} and \ref{tab:results-metrics_1}. Fortunately, some authors independently verified distinct profitability levels when comparing assets of different types of \cite{Ye2020} or from various countries \cite{Eilers2014, Park2019, Wang2019a}. For instance, \cite{Wang2019a} observed better performance on data from emerging countries’ markets and attributed these results to the system’s capacity to explore the vulnerability to speculative operations of emerging markets. According to \cite{WangGang2012}, FOREX markets’ efficiency varies depending on the currency pair selection. Considering this conclusion, a review of works about market prediction using text mining \cite{KhadjehNassirtoussi2014} hypothesized that these pairs’ choices might affect the predictability results. Thus, in the study by \cite{Ye2020}, we believe choosing a less mature currency for trading (cryptocurrency) might have led to notably better profitability than using high-capital stocks.

Even though profitability evaluation is essential, additional aspects regarding transaction costs (TC) and comparison to benchmarks can also be valuable to strengthen the significance of findings and assess proposed methods' applicability in real-world situations. Hence, in both Table \ref{tab:results-trading} for single asset trading and Table \ref{tab:results-portfolio} for portfolio optimization, we list these aspects and other information concerning findings, contributions, or novelty to the RL body of work on financial tasks. Transaction costs -- which can assume a percentage \cite{Weng2020} or fixed values \cite{Gabrielsson2015} -- are essential when simulating realistic scenarios, given they are present in most markets and may incur diminishing returns if an investor makes too many operations. Unsurprisingly, only five studies did not use any TC, while three authors did not specify adopted values \cite{Gabrielsson2015, Si2017, Jeong2019}. To manage costs, some authors \cite{Almahdi2017, Almahdi2019} proposed stop-loss mechanisms that made their architecture resistant to TC. Also, \cite{DengYue2017} and \cite{LiYang2019} showed RL methods outperforming supervised learning prediction strategies by adapting to TC.

\noindent\begin{minipage}{\textwidth}
{\small
\setlength{\LTleft}{-20cm plus -1fill}
\setlength{\LTright}{\LTleft}
\begin{longtable}{@{} p{0.08\textwidth} p{0.075\textwidth} p{0.2\textwidth} p{0.3\textwidth} p{0.35\textwidth}  @{} }
\caption{Comparison of single asset trading problem regarding benchmark strategies, alternative approaches settings, transaction cost (TC), and relevant findings, contributions or novelty to RL body of work on financial tasks.}
\label{tab:results-trading}\\
\toprule
\multicolumn{5}{@{} p{\linewidth}}{\textbf{Single Asset Trading}} \\
\toprule
\textbf{Reference} & \textbf{TC} & \textbf{Benchmarks} & \textbf{Alternative approaches} & \textbf{Findings, Contributions or Novelty} \\ \hline

\cite{Zarkias2019} &
  30\% over return &
  No &
  DDQN without trail &
  Price trailing as a new formulation for the problem \\ \hline
\cite{Ponomarev2019} &
  Fixed: 2.5 units &
  No &
  Various neural network architectures &
  Explored various A3C architectures to suggest best design choices \\ \hline
\cite{Jeong2019} &
  No &
  BH &
  RL using index, using correlation or neural networks to select the stocks &
  Isolation of one DQN to find the amount to be traded and another to decide the buy, sell, or hold action \\ \hline
\cite{Wu2019} & 
  Not Spec. & 
  No & 
  1st: FC ANN, long short-term memory networ \newline 2nd: 4 varied combinations of tech. indicators as features & 
  Examined combinations of technical indicators for states and identified best one \\ \hline
\cite{LiYang2019} & 
  0.01\%, 0.05\%, 0.04 unit &
  BH, Supervised Learning methods with fixed rules and no costs &
  A3C or double DQN + stacked denoising autoencoders &
  - RL outperformed prediction Supervised Learning \newline - Verified benefits of denoising techniques \newline - A3C worked best than double DQN \\ \hline
\cite{GiacomazziDantas2018} & 
  No & 
  BH, other 2 Financial strategies & 
  No & 
  Design of custom reward for contemplating won/lost opportunity \\ \hline
\cite{Alimoradi2018} & 
  0.75\% & 
  BH, genetic network programming method with rule accumulation heuristic & 
  No & 
  - Backward a combination of Sarsa learning and Q-learning \newline - Combined RL with genetic metaheuritics \\ \hline
\cite{Carapuco2018} & 
  Not Spec. & 
  No & 
  No & 
  Managed on stabilizing training so it came close to test results \\ \hline
\cite{Spooner2018} & 
  No & 
  Financial Strategy, Sup. Learn. method, 5 different RL techniques & 
  RRL with variations of: reward dampening, state encoding & 
  - Linear combination of tile codings (LCTC) encoding works very well for state representation and can increase stability \newline - Best reward: Asymmetrically dampened reward \\ \hline
\cite{DengYue2017} & 
  Fix.: 1\(\sim\)2 un. \newline Global: 0.1\% & 
  BH, deep reinforcement learning, SCOT, 3 Prediction Supervised Learning models & 
  - Main: DDR with fuzzy encoding and various rewards (SR, total profit) \newline - Global Market: fuzzy DDR & 
  - Introduction of fuzzy logic and expanded past actions for state representation \newline - System outperformed fixed strategy fixed rules prediction systems \\ \hline
\cite{Si2017} &
  Fix.: 1 &
  BH, RRL &
  No &
  - Best reward: Multi-objective of max. profit and SR \\ \hline
\cite{Feuerriegel2016} &
  0.1\% &
  BH, Momentum Trading &
  Supervised Learning (Random Forest), Rule-based sentiment trading &
  Proposed combination of RL methods with sentiment analysis of news data sources \\ \hline
\cite{Gabrielsson2015} &
  Fix.: 0, 0.5 pnt &
  BH, Random, RRL &
  No &
  Combination of RRL with lagged candlestick features \\ \hline
\cite{Eilers2014} &
  No &
  BH, Static Rule Strategy &
  No &
  - Embedded special days information into RL and outperformed static strategies and BH. \newline - Explored impact of events such turn-of-the-month. \\ \hline
\cite{ZhangJin2014} &
  0.30\% &
  BH, Random &
  Elitist, Avg. Elitist, Two Versions of Multiple Elitist &
  - Selection of the best internal weights of RL model \newline - Avg. elitist worked little better than multiple elitist \\ \hline
\cite{Maringer2014} &
  0.30\% &
  BH, Random &
  Indicators variations as part of state for RS-RRL: conditional and implied volatility, RSI, trading volume &
  RS-RRL with volume indicator in state rep. was the best variation \\

\bottomrule
\end{longtable}}
\end{minipage}

Regarding benchmarks, \cite{Wang2019a} and \cite{Ye2020} were the only surveyed authors to adopt a method of another work here presented (\cite{DengYue2017} and \cite{JiangZhengyao2017}, respectively), as a benchmark. Alternatively, we observe distinct benchmark preferences among supervised and reinforcement learning models, heuristics, classic financial strategies, indicators-based (technical or fundamental) strategies, and others. Despite this, there is a notable predominance of the \textit{buy and hold} (BH) strategy -- present in eleven single asset articles (68.75\%) and ten portfolio optimization articles (76.92\%) -- also known in portfolio optimization problems as equal weights or uniform BH (EW-BH or U-BH). This prevalence is not a surprise since BH is a simple and effective passive financial strategy. Investors buy and hold an constant amount of an asset in time (single asset problem) -- or a constant group of assets in time (portfolio problem) -- expecting in the long-term, despite short-term fluctuations, assets’ value will ultimately increase. Notably, although most studies reported better profitability than selected benchmarks, not all methods outperformed BH \cite{Maringer2014, GiacomazziDantas2018} or other benchmarks \cite{JiangZhengyao2017}.

\noindent\begin{minipage}{\textwidth}
{\small
\setlength{\LTleft}{-20cm plus -1fill}
\setlength{\LTright}{\LTleft}
\begin{longtable}{@{} p{0.08\textwidth} p{0.05\textwidth} p{0.2\textwidth} p{0.375\textwidth} p{0.35\textwidth}  @{} }
\caption{Comparison of Portfolio optimization problem regarding benchmark strategies, alternative approaches or settings, transaction cost (TC), and relevant findings, contributions or novelty to RL body of work on financial tasks.}
\label{tab:results-portfolio}\\
\toprule
\multicolumn{5}{@{} p{\linewidth}}{\textbf{Portfolio optimization}} \\
\toprule
\textbf{Reference} & \textbf{TC} & \textbf{Benchmarks} & \textbf{Alternative approaches} & \textbf{Findings, Contributions or Novelty} \\ \hline
\endfirsthead
\endhead
\cite{Weng2020} & 
  0.25\% & 
  Constantly rebalanced portfolio (CRP), other 2 financial strategies & 
  CNN &
  Uses XGBoost for the feature selection and CNN for state feature extraction using three-dimension attention networks. \\ \hline
\cite{Aboussalah2020} & 
  0.55\% & 
  EW-BH, Mean-Variance Optimization (MVO), Risk Parity & 
  Hyperparameters combinations &
  Time Recurrent Decomposition stacking RRL structures updating recent decisions. \\ \hline
\cite{Ye2020} & 
  0.25\% & 
  Equal Weights BH (EW-BH), CRP, other 2 financial strategies & 
  DPM \cite{JiangZhengyao2017} &
  - Textual news word embedding (GloVe) composition of state \newline - DDPG method \\ \hline
\cite{Park2019} & 
  0.25\% & 
  EW-BH, Random, other 2 financial strategies & 
  No &
  Different state composition together with encoding reduction \\ \hline
\cite{Almahdi2019} & 
  0.1\%, 0.15\%, 0.2\%, 0.25\% & 
  Equal Weights BH (EW-BH), particle swarm optimization heuristics & 
  RRL + particle swarm optimization and variations&
  - Calmar ratio as heuristic fitness function and reward function \newline - RRL combinations outperformed their raw particle swarm optimization counterparts \\ \hline
\cite{Wang2019a} &
  0.10\% &
  EW-BH, other 3 financial strategies, fuzzy deep direct reinforcement method (DDR) \cite{DengYue2017} &
  RL and each feature extraction/representation attention network method &
  - Addresses: risk, profitability and stability of model; interrelationship between assets; model behavior explanation \newline - Adaptation of the RL framework to reach a buy-winners-and-sell-losers (BWSL) design \\ \hline
\cite{Yu2019} & 
  0.20\% & 
  Constantly rebalanced portfolio (CRP) & 
  8 Different combination of proposed Modules &
  - Prediction submodule adoption is beneficial \newline - Another work that used expert imitation \\ \hline
\cite{Pendharkar2018} & 
  No & 
  Equal Weights BH (EW-BH), Theoretical Ceiling. & 
  Methods: Q-learning vs SARSA; Actions: continuous vs discrete; Rewards: Return vs DSR (only for discrete actions); Learning: Adaptive vs Static; Trading Frequency & 
  Sarsa with continuous actions, adaptive learning and lower trading frequency were the best options \\ \hline
\cite{Kang2018} & 
  Not Spec. & 
  EW-BH & 
  Single Thread A3C vs Multi Thread A3C (A3C is selected) &
  - Multi-threaded is more stable than single-thread \newline - Model lost to S\&P 500 index \\ \hline
\cite{DingYi2018} & 
  No & 
  No & 
  Different training reward functions: AR, Max Drawdown, SR, Exceed Return, win rate &
  - Best reward function was Max Drawdown \newline - Incorporated imitating investor behavior \\ \hline
\cite{Almahdi2017} & 
  0.1\%, 0.15\%, 0.2\%, 0.25\% & 
  5 variations of portfolio BH, Hedge Funds & 
  Rewards variations: Sterling Ratio, SR, Calmar Ratio; Actions variations Equal Weight (EW), Variable Weight (VW) & 
  - Showed the calmar Ratio as best reward and VW as best for actions \newline - Introduced stop-loss mechanism to deal with costs \\ \hline
\cite{JiangZhengyao2017} & 
  0.25\% & 
  EW-BH, CRP, other, 2 financial strategies, 2 SL & 
  No &
  Lost to PAMR in returns and to Online Newton Step in SR \\ \hline
\cite{Zhang2016} & 
  0.30\% & 
  Equal Weights BH (EW-BH), Random & 
  GA-RRL variations of selected indicators for state representation &
  - Use of GA metaheuristic for selecting state representation. \newline - Only variation that did not use indicators had a worst performance. \\

\hline
\end{longtable}}
\end{minipage}

To summarize and highlight reviewed articles' achievements, we separate the comparisons and contributions present in Tables \ref{tab:results-trading} and \ref{tab:results-portfolio} according to some of the main design aspects (depicted in Fig. \ref{fig:pipeline}) that we examined. Subsequently, using some of the workflow pipeline divisions, we list a few of the achievements:

\begin{description}

  \item \textbf{Preprocessing and feature representation:} \cite{Spooner2018} found it advantageous to adopt a preprocessing technique (i.e., the linear combination of tile codings (LCTC)) instead of directly using technical indicators in the state composition. Similarly, \cite{LiYang2019} highlighted the improvement of running both the input prices and technical indicators through a stacked denoising autoencoders layer instead of directly in the state. Also, the stacked denoising autoencoders adoption proved more influential over results than the choice of an RL method (A3C or DQN). For better noise reduction over price data, \cite{DengYue2017} examined feature representation schemes and found that using an initial fuzzy layer improved AR and SR. Alternatively, \cite{Wang2019a} experimented with different attention network mechanisms for learning feature representations in a long short-term memory model. \cite{Wang2019a} observed better results when the proposed attention mechanism encompassed the correlation of different stock trends followed by a priori stock interrelationship information incorporation. In this study, the proposed AlphaStock method outperformed the fuzzy deep direct reinforcement (DDR) benchmark \cite{DengYue2017}, which was the best benchmark. As such, future work could benefit from exploring a combination of both fuzzy and attention techniques.
  
  \item \textbf{State composition:} \cite{Wu2019} evaluated the combinations of a diverse number of technical indicators as features for state representation. Moreover, \cite{Wu2019} found that a specific combination of 31 indicators got more profit than other combinations with more (e.g., 35) or fewer indicators (e.g., 10). \cite{Zhang2016} compared variations of the proposed genetic algorithm (GA) meta-heuristic scheme for selecting an appropriate set of technical indicators for state representation. Adopting the technical indicators, the meta-heuristic models could consistently outperform random trading and the models using only price time-series as state representation. \cite{Maringer2014} explored technical indicators used individually as inputs for the proposed regime-switching RRL method (RS-RRL) and verified that the trading volume indicator led to higher SR and statistical confidence. Two studies introduced a promising new type of fundamental indicator for state composition; They extracted information from news using \textit{natural language processing} (NLP) techniques such as \textit{sentiment analysis} \cite{Feuerriegel2016} or \textit{word embeddings} \cite{Ye2020}.
  
  \item \textbf{RL Methods:} \cite{Pendharkar2018} pointed to SARSA consistently outperforming Q-learning in all equivalent architecture configurations. \cite{LiYang2019} implemented a DQN and an A3C with experiments indicating better performance in AR and SR by the A3C model. Besides, \cite{Kang2018} confirmed the benefits of a multi-threaded A3C version over a single thread. \cite{Wu2019} switched between two supervised models as the inner core of an RL approach. They concluded that a long-term memory network achieved higher profitability than a fully connected deep neural network. \cite{Yu2019} proposed three modules (infused prediction module, data augmentation module, behavior cloning module), associated them in seven possible combinations, and assessed each one's incremental gains. The infused prediction module was the most important in improving AR and Max Drawdown metrics, while the data augmentation module reduced overfitting. The behavior cloning module lowered portfolio volatility and Max Drawdown. Finally, the consolidated architecture used all three modules and outperformed the other combinations in four risk-adjusted metrics, including SR and Sortino ratio.
  
  \item \textbf{Actions:} \cite{Pendharkar2018} experimented with both continuous (portfolio weight vector) and discrete actions (predetermined portfolio proportions). Besides, for discrete actions, \cite{Pendharkar2018} compared total return and Differential Sharpe Ratio (DSR) rewards, with DSR leading to better results. In these experiments, an exhaustive comparison showed that continuous action methods consistently outperformed their discrete action counterparts. For the portfolio optimization task, \cite{Almahdi2017} pointed out that with equal weight (EW) actions, forcing the rebalance of stocks' weight to an equal distribution seems prejudicial. Thus, variable weight (VW) actions produced better AR and SR metrics than equal weight (EW) actions.
  
  \item \textbf{Rewards:} Even though \cite{DengYue2017} found similar overall results between optimizing for the total profit or SR, the authors suggested SR adoption as it allowed better consistency in balancing profits with risk. In contrast, \cite{DingYi2018} identified that adopting either AR or Max Drawdown reward functions as the models' objective function yielded better results than using the SR. \cite{Almahdi2017} explored SR and Max Drawdown-based rewards (Calmar and Sterling ratios). With transaction costs included, there were inferior results when using SR as the reward, so this method's version was lost to BH. Interestingly, as the transaction costs increased, the Calmar ratio went from a slightly better reward option to an outstanding choice. \cite{Spooner2018} proposed a mechanism for dampening rewards (i.e., reducing) that diminished the model's actions' volatility and tendency to take speculative actions, thus making it more stable and profitable. Additionally, dampening profits and losses was less effective than dampening only profits (asymmetric dampening). Instead of searching for a single adequate reward function, \cite{Si2017} multi-objective approach explained in Section \ref{subsec:actions_rewards}, outperformed benchmarks in total profit and SR.

  \item \textbf{Other approaches:} A fair number of authors adopted heuristic or meta-heuristic methods. \cite{Almahdi2019} investigated three different varieties of the particle swarm optimization meta-heuristic. It concluded that particle swarm optimization was the most appropriate for combination with a base RRL model for the many optimization liaisons. \cite{Maringer2014} assessed the benefits of the \textit{differential evolution} heuristic in enhancing the regime-switching (RS) capabilities of the RS-RRL model. \cite{Alimoradi2018} showed the highest AR average result by using the \textit{league championship algorithm} (LCA) meta-heuristic to select the most fitted strategies trained by the RL method. Among other approaches, \cite{DingYi2018} and \cite{Yu2019} pointed out the benefits of learning strategies by mimicking or imitating human investors' behavior. \cite{Carapuco2018} proposed a time-skipping technique that guaranteed diversification of data exposure throughout training and evaluation that increased test results confidence. \cite{Wang2019a} evaluated model interpretability through a proposed metric for checking the influence of each feature over the asset selection and understanding the learning agent's behavior better.

\end{description}


\section{Discussion}
\label{sec:discussion}

From the reviewed papers' observed characteristics, we propose some relevant discussions, and we produce some suggestions for future works. This section aims to highlight common aspects of papers and give insights into future research opportunities. Before starting the discussion, we emphasize essential reinforcement learning (RL) characteristics. Most RL methods are built upon theoretical constructs rooted in solid mathematical ground. Therefore, by exploring the coherence between the problem formulation and the RL fundaments, we advise several topics of discussion to expose the knowledge frontier. 

Also, there are recommendations about avoidance of pitfalls like \textit{the deadly triad}, or the \textit{curse of dimensionality} \cite{Sutton2018}, which are related to issues regarding convergence and stability.

In this section, we show the versatility and flexibility of RL methods by discussing the diverse directions taken by surveyed studies and exhibiting different future potential paths. We point to how some authors have been alienated from discussing relevant RL fundamentals. Also, we argue that these essential concepts are necessary for identifying gaps and research opportunities. Furthermore, it is necessary never to lose track of theoretical background because it can help design and apply better expert systems. Moreover, apart from these constructs, RL provides much room for framing a problem and designing a solution.

\subsection{Contextual bandit and full reinforcement learning formulations}

In Section \ref{subsec:preprocessing}, we examined works under the theoretical viewpoint of the formulation of problems as a contextual bandit or a full reinforcement learning problem (as described in Section \ref{subsec:contextual_bandit}). First, we observed that, in most cases, these concepts seem overlooked when designing the system. As the reviewed works are technically sound, it is unclear if this information was left out purposely or inadvertently. However, as we discussed in the mentioned section, differently from game domains, the formulation of RL problems is not as straightforward for the financial market domain. For this specific characteristic, we suggest that authors explicitly mention the problem formulation they aim at when designing their architectures and which assumptions are made to fit the problem. Thus, because these conceptual aspects can directly impact the design and implementation of methods, future papers should be more attentive to these and other theoretical details.

Moreover, while all analyzed studies used conventional RL methods -- generally designed for the typical RL formulation -- there are alternative solutions commonly associated with contextual bandit problems, such as the upper confidence bound (UCB) algorithm. However, all reviewed contextual bandit works opted for using typical RL methods with some adjustments, such as the reward function. Although there is nothing conceptually wrong with using adjusted RL solutions for the contextual bandit framing, it is logically conceivable that better results (or at least similar results in a more efficient way) could be achieved using methods designed for this type of formulation. Hence, future approaches that employ the contextual bandit problem formulation may consider exploring bandit methods and comparing it to the most popular RL solutions. The work by \cite{LihongLi2010} -- described in Section \ref{subsec:contextual_bandit} -- could serve as inspiration for future articles as an effective and elegant way of framing a problem in a contextual bandit formulation and then applying a bandit method.

According to the main RL literature \cite{Sutton2018}, contextual bandit formulations tend to lead to greedy solutions, which is confirmed to be a problem in the web marketing domain. Consequently, answering this research question would make it easier for future works to concentrate their efforts more efficiently. In Section \ref{subsec:contextual_bandit}, we mentioned the comparison executed by \cite{Theocharous2015}, who established the benefits of using a full RL formulation and a full RL method instead of employing a bandit method in a contextual bandit formulation. Likewise, similar comparisons could lead to similar conclusions and help address this crucial question of which way of framing a financial trading problem is the most promising.

Additionally, researchers should feel encouraged to adopt the abstraction proposed by \cite{Spooner2018} -- which we mentioned in Section \ref{subsec:preprocessing} -- that conceptually separates the complete representation of the state into market-states and agent-states. This abstract separation may help with the state representation's design, tweaking, and fine-tuning. Besides, there is an excellent opportunity to explore abstractions similarly, such as in \cite{Wu2019} (Section \ref{subsec:results}). Such comparative works could help establish a reference about which features to use and the adequate number of features for each state type.

\subsection{Markov decision process design}

We showed that many papers successfully explored technical indicators in preprocessing for state representation. Furthermore, we found that using deep learning encoders also yielded promising noise reduction and feature extraction results. Moreover, as discussed in Section \ref{subsec:results}, there are benefits in coupling both encoders and technical indicators in preprocessing. In this regard, most authors could explore this direction and employ a combination of deep learning encoders and just a few of the most common technical indicators reported in Section \ref{subsec:preprocessing}. Besides, there is still room to spend some effort investigating other types of encoders to improve feature extraction and denoising and other types of neural network architectures to take advantage of the non-linear approximation property to improve feature extraction and denoising.

In Section \ref{subsec:results}, we identified that just two works \cite{Almahdi2017, Pendharkar2018} evaluated the influence of different action designs on the results. Consequently, as this aspect has not yet been examined much and seems to impact the system's design significantly, it could be further investigated to expand its knowledge.

We mentioned in Section \ref{subsec:actions_rewards} that optimization is performed towards a single reward objective for most methods. For these single reward optimization methods, authors mostly examined SR and/or AR as a reward and then used the Max Drawdown only as an evaluation metric and not as a reward function. Nevertheless, when some works compared the use of AR or SR against Max Drawdown-based reward functions such as the Calmar ratio, the AR and SR indicators were outperformed. Hence, while further investigative comparisons are still required, the current body of work supports an increase in the adoption of the Calmar ratio and other Max Drawdown-based measurements as reward functions for future research. The elaboration of customized rewards \cite{Spooner2018, GiacomazziDantas2018} also tends to work well, but no paper compared these types against the most typical rewards, so there is no indication about in which circumstances they might be more suited. Thus, it would be appropriate that future research employing customization of rewards includes comparisons of the proposed reward function against typical rewards. Additionally, as observed in Section \ref{subsec:results}, only one approach \cite{Si2017} successfully explored multi-objective methods as an alternative to the majority of single reward methods. Therefore, we believe that authors should feel encouraged to work towards growing the body of research about multi-objective rewards.

In Section \ref{subsec:results}, overall results pointed to the enhancement of RL methods through the use of heuristic or meta-heuristic techniques. Moreover, we observed diverse ways of achieving these improvements, such as using genetic algorithms to select adequate indicators to represent a state \cite{Zhang2016}. We also examined that \cite{Alimoradi2018} used meta-heuristics to select an appropriate trading RL policy among trained RL policies. In this sense, employing heuristic methods is a great way to expand the number of possible RL system designs. Thus, we strongly suggest that future research continue to explore these promising combinations.

We discussed in Section \ref{subsec:rl_methods} diverse ways employed to deal with the exploration-exploitation dilemma. The use of \(\varepsilon\)-greedy mechanism was the most prevalent among TD works, and the use of stochastic policies amid policy approximation works tackled the portfolio optimization tasks. 
We also discussed in Section \ref{subsec:rl_methods} that a noisy environment helps to improve exploration in deterministic RL methods. However, future research would be illuminated by examining how the methods of stochastic policy approximation compare to their deterministic counterparts in single asset tasks.

As presented in Section \ref{subsec:preprocessing}, a reasonable amount of articles successfully combined technical and fundamental indicators to compose RL states. In particular, two works have shown encouraging results by adopting natural language processing (NLP) techniques, such as sentiment analysis \cite{Feuerriegel2016} and word embeddings \cite{Ye2020}, to incorporate textual information from the news. Additionally, these works confirmed that, even though the NLP field of research is vast, the early-mentioned flexibility of RL methods allows for straightforward yet powerful combinations. This conjunction of NLP with RL is auspicious, not much investigated, and its exploration could be vital in producing research breakthroughs. Researchers may present ways to improve on the previously mentioned methods as a starting point. Also, researchers could enhance the \cite{YangSteveY2018} architecture by adopting strategy learning instead of predetermined rules.

\subsection{Reinforcement learning methods}

Our analysis showed that even in recent years, the recurrent reinforcement learning (RRL) method for financial tasks is still a standard for our study field. We also observed that employing the RRL base model or its extensions is usual as a standard. Consequently, this consolidation of RRL, as the most used method for financial tasks, happened due to its simplicity, effectiveness, and flexibility for extensions. In this regard, as discussed in Section \ref{subsec:results}, the community continues to find successful ways to extend the base method with the introduction of modern deep learning structures, incorporation of technical indicators, deep learning encoding, and other techniques. Hence, there are still many opportunities to expand the body of work around RRL approaches.

We also observed that, recently, among the surveyed works, it appears to be an increase in actor-critic methods. Furthermore, similar to RRL methods, actor-critic methods also possess some flexibility to extensions, but in this latter case, due to the use of the baseline functions mechanism (discussed in Section \ref{subsec:rl_methods}). The advantage function was the only baseline function used among reviewed works, and therefore, we believe that researchers should feel encouraged to investigate other baseline functions. Moreover, these methods are promising since, as addressed in Section \ref{subsec:results}, the study in \cite{LiYang2019} showed that an actor-critic A3C method outperformed a critic DQN. However, we surveyed no work that compared an actor-critic method to any modern extension of RRL. In this sense, future researchers should feel encouraged to compare actor-critic methods against the more consolidated RRL approaches. These comparisons could increase confidence in actor-critic methods and open new research directions.

\subsection{Considerations over the Markov property}

In the single asset trading or portfolio optimization problem, it is possible to infer that the most realistic modeling is not the Markov Decision Process due to unobserved information that explains the actual state. To satisfy Markov's property, we would need a state that summarizes all the information necessary to make predictions of future states \cite{Sutton2018}. In financial time-series, having a state that is a compact history summary becomes impracticable since history grows and there is no evident recurrence. The reviewed papers do not address the Markov process matter in their work and what hypothesis they are assuming, except for \cite{LiYang2019}.

The Partially Observed Markov Decision Process (POMDP) is an approach in which we assume latent states that produce the environment's observations. The distribution over the latent space gives the belief state assumed to be the Markov state. Another approach is using Predictive State Representations, in which we define the Markov state as a d-vector of the probabilities \cite{James2004}. This vector is updated by a state-update function that looks into the future, not the records. A third approach could be using a function to approximate the state to the latest observations, indirectly employed by some reviewed articles when applying the preprocessing phase. Using technical indicators, forecasting methods, and non-linear encoders, the authors bring the state closer to the Markov state.

When the Markov property is only approximately satisfied, long-term predictions can be degraded and inaccurate. Short-term and long-term approximation objective has no theoretical guarantees \cite{Sutton2018}. The problems associated with the short and long-term predictions are related to the degradation of the RL methods' performance over time as the financial time-series could change. The model only has the beliefs or function approximation over the actual Markov states. By considering POMDP modeling, we could apply different techniques following the RL trend for POMDP, opening alternative paths to explore RL applied to trade.

\subsection{Using synthetic time-series}

Most of the reviewed papers did not use synthetic time-series for method comparison. The synthetic time-series could be an excellent alternative to investigate the method advantages in specific scenarios that their construction is based on mathematical functions. The approach used by \cite{Yu2019} with a GAN (Generative Adversarial Network) in the data augmentation module is essential because of two aspects:
1) The augmentation of data for the model training improves generalization for the policy estimator; 
2) The use of synthetic time-series gives a reference since they have an original function behind them and are not entirely random.

\subsection{General standards}
\label{subsec:general_standards}

In Section \ref{subsec:raw_data}, we verified a lack of standards or benchmarks when gathering data for a trading problem. There are no consistent criteria across works for gathering data, and each group of researchers has preferences about data characteristics such as country, amount, period, time-frame. In this sense, most papers tackle different tasks of the same research problem domain. \cite{KhadjehNassirtoussi2014} detected a similar issue in the articles about sentiment analysis techniques applied to the financial domain. Therefore, standards are a general predicament in the body of work that uses machine learning techniques for financial domain problems.

Moreover, \cite{KhadjehNassirtoussi2014} suggested that future works in that area of study should address this lack of standardization. Some years later, there was an effort to address this problem with the launch of the SemEval 2017 Task 5 competition \cite{Davis2016, Cortis2017}. This competition generated a gold standard financial dataset. The research area explored here would greatly benefit from a similar effort.

In Section \ref{subsec:results}, we highlighted two main implications that this lack of standardization had on the overall analysis of results in Section \ref{subsec:results}: (i) general profitability in very distinct scenarios suggests broad market exploitability; (ii) particular market exploitability can not be guaranteed. Hence, combining these two observations justifies the elaboration of not just one standard financial dataset but several datasets for different tasks. The task of trading an asset in, for example, a high-frequency intraday scenario is very different from doing it for a long-term objective, hence the need for this dataset variety. Consequently, researchers should expect to see different solutions having various results depending on each task.

The training and evaluation of methods are intrinsically related to the characteristics of the gathered data. Thus, it is not surprising that we found another intriguing aspect similar to the work by \cite{KhadjehNassirtoussi2014}: few analyzed studies used the rolling window setup. In Section \ref{subsec:setups}, we considered the various benefits of the rolling window setup over the static window. Additionally, we have another essential remark concerning the rolling window setup and the generalization, stability, and convergence issues in RL methods. The rolling window setup could help check these issues, ensure consistency of results, and facilitate the evaluation of statistical significance. Consequently, we believe future works should pay more attention to the rolling window setup and consider making this the standard setup method.

\cite{Spooner2018} were the only authors to make their source codes publicly available. Hereafter, more authors must adopt this good practice as a norm to increase replicability and confidence in results. Moreover, this convention could make it more likely and suitable for researchers to improve upon previous architectures.

Market efficiency is a fundamental topic in the financial domain, as mentioned in several parts of this paper. Moreover, there are articles in the financial domain about indices for measuring market efficiency \cite{Tran2019}. Hence, with the adoption of better standardization practices, researchers might be able to open a new research field in devising market efficiency metrics based on RL methods.

Although RL is a field with very well established literature \cite{Sutton2018} that guarantees consistency in the formalism and terminology employed across works \cite{Silver2017, Mnih2016}, we observe that this standard was not always present in the surveyed papers. Therefore, it would contribute to cohesion and organization for the body of work if future research follows the primary RL literature’s overall formalism tenets.


\section{Conclusion}
\label{sec:conclusion}

In this work, we executed an exhaustive review of the past six years of the literature regarding the Reinforcement Learning framework for solving financial asset trading tasks. We believe our survey is one of the first works to present such a systematic examination while adopting approaches rooted in the RL concepts. This approach allowed us to propose a workflow pipeline representation of the typical elements among the reviewed articles' systems design. The pipeline representation aided in achieving the following main contributions: First, this representation supported a systematic investigation of papers and employed techniques according to each design element. Second, it enabled us to identify state-of-the-art methods amidst the increasing trend of RL methods for trading tasks. Third, it helped us cover the integration of RL theoretical concepts with the trading problem domain and advert the importance of the precise definition of assumptions taken by the authors in light of the RL tenets. Finally, we suggested different approaches that we judged promising and could offer a firm ground for future research efforts.

Integrating many machine learning techniques with RL representation of the problem leads to a hopeful future. From the results presented by the reviewed papers, we can also observe the increase in articles adopting modern RL approaches, such as actor-critics. Also, the developments in natural language processing, forecasting methods, encoders, and many other areas can increase the RL agent's information, improving the decision support's assertiveness.

In this work, we hoped to highlight potential gaps and provide insightful knowledge to promote advances in financial trading solutions. Furthermore, given the high expectations that might exist over recent RL achievements in other domains, we aimed to give a more realistic view of RL techniques' capabilities in the trading domain. Because of the lack of standardization, the numerical comparison between works is harmed; therefore, authors should pursue a common standard. Moreover, upcoming works may find it advantageous to explore and extend the discussion to develop alternative paths over the enormous variety of ideas combined.

\section*{Credit author statement}

\textbf{Leonardo Kanashiro Felizardo}: Conceptualization, Methodology, Formal analysis, Investigation, Writing - Original Draft, Writing - Review \& Editing, Project administration \textbf{Francisco Caio Lima Paiva}: Conceptualization, Methodology, Formal analysis, Investigation, Writing - Original Draft, Writing - Review \& Editing  \textbf{Anna Helena Reali Costa}: Conceptualization, Methodology, Validation, Writing - Review \& Editing, Supervision, Project administration, Funding acquisition \textbf{Emilio Del-Moral-Hernandez}: Writing - Review \& Editing, Supervision, Funding acquisition.

\section*{Acknowledgements}
This study was financed in part by the  \textit{Coordenação de Aperfeiçoamento de Pessoal de Nível Superior} (CAPES Finance Code 001), Brazil, and \textit{Conselho Nacional de Desenvolvimento Científico e Tecnológico} (CNPq) grants 310085/2020-9 and 88882.333380/2019-01 and by \textit{Ita\'{u} Unibanco S.A.} through the \textit{Programa de Bolsas Ita\'{u}} (PBI) program of the \textit{Centro de Ci\^{e}ncia de Dados} (C$^2$D) of \textit{Escola Polit\'{e}cnica} of USP.


\bibliographystyle{plain}

\bibliography{Bibliography_RS}



\end{document}